\documentclass[letterpaper, 10 pt, journal, twoside]{IEEEtran}
\usepackage{amsmath}
\DeclareMathOperator{\st}{s.t.}
\usepackage{amssymb}    
\usepackage{mathtools} 
\usepackage{siunitx}
\usepackage{graphicx}
\usepackage{subfigure}
\usepackage[ruled,vlined]{algorithm2e}
\usepackage{tabularx}
\usepackage{enumitem}
\usepackage{balance}
\usepackage{tabularx}
\usepackage{booktabs}
\usepackage{fancyhdr}

\newcolumntype{Y}{>{\centering\arraybackslash}X}
\usepackage[hidelinks, colorlinks = true, linkcolor = red, citecolor = green]{hyperref}
\usepackage{xcolor}
\newcommand{\T}{^\mathsf{T}}
\IEEEoverridecommandlockouts

\fancypagestyle{firstpage}
{

    \fancyhead[L]{}    
    \fancyhead[R]{}
    \fancyhead[C]{\textcolor{gray}{This paper has been accepted for publication at Robotics: Science and Systems, 2024. \\ Please cite the paper as: J. Huang, Y. Tang, and K. W. Samuel Au, \\ ``Homotopic path set planning for robot manipulation and navigation," Robotics: Science and Systems, 2024.}}
    \fancyfoot[]{}
}
\fancypagestyle{RSShead}
{

    \fancyhead[L]{\textbf{\textsf{Robotics: Science and Systems 2024 \\ Delft, Netherlands, July 15-July 19, 2024}}}    
    \fancyhead[R]{}
    \fancyhead[C]{}
    \fancyfoot[]{}
}

\begin{document}
\title{
Homotopic Path Set Planning for Robot Manipulation and Navigation}
\author{
        Jing Huang$^{1, 2}$, Yunxi Tang$^1$, and Kwok Wai Samuel Au$^{1,2}$ \\
        $^1$Department of Mechanical and Automation Engineering, The Chinese University of Hong Kong \\
        $^2$Multi-Scale Medical Robotics Center, Hong Kong SAR \\
        Email: \{huangjing, yxtang\}@mae.cuhk.edu.hk, samuelau@cuhk.edu.hk
}
\maketitle
\thispagestyle{firstpage}

\begin{abstract}
This paper addresses path set planning that yields important applications in robot manipulation and navigation such as path generation for deformable object keypoints and swarms. A path set refers to the collection of finite agent paths to represent the overall spatial path of a group of keypoints or a swarm, whose collective properties meet spatial and topological constraints. As opposed to planning a single path, simultaneously planning multiple paths with constraints poses nontrivial challenges in complex environments. This paper presents a systematic planning pipeline for homotopic path sets, a widely applicable path set class in robotics. An extended visibility check condition is first proposed to attain a sparse passage distribution amidst dense obstacles. Passage-aware optimal path planning compatible with sampling-based planners is then designed for single path planning with adjustable costs. Large accessible free space for path set accommodation can be achieved by the planned path while having a sufficiently short path length. After specifying the homotopic properties of path sets, path set generation based on deformable path transfer is proposed in an efficient centralized manner. The effectiveness of these methods is validated by extensive simulated and experimental results.
\end{abstract}
\IEEEpeerreviewmaketitle

\section{Introduction}
Simultaneously generating paths for multiple agents finds vast applications in robotics. For instance, planning multiple robots' paths with certain collective properties has been extensively studied in multi-robot navigation tasks like surveillance, formation flight, and collaborative transport \cite{J. Alonso-Mora 2017}-\cite{W. Hönig 2018}. Manipulative tasks also demonstrate similar needs. Particularly, robotic deformable object manipulation (DOM) remains challenging today. One core reason is that widely populated constrained environments in reality make DOM far more complicated beyond pure control approaches \cite{J. Sanchez 2018, V. E. Arriola-Rios 2020} and entail planning \cite{P. Jimenez 2012, H. Yin 2021}. The reliance on models or simulations renders deformation planning not easily applicable. An effective alternative is directly planning spatial paths for deformable objects (DOs). Considering DO keypoints provides a tractable way to depict the object states \cite{J. Huang 2023 TRO}. Analogous to multi-robot paths, keypoint paths should be specified coordinately. For consistency, a path set here refers to the collection of paths for a robot team or object keypoints.

\begin{figure}[t]
    \minipage{1 \columnwidth}
     \centering
     \includegraphics[width= 1 \columnwidth]{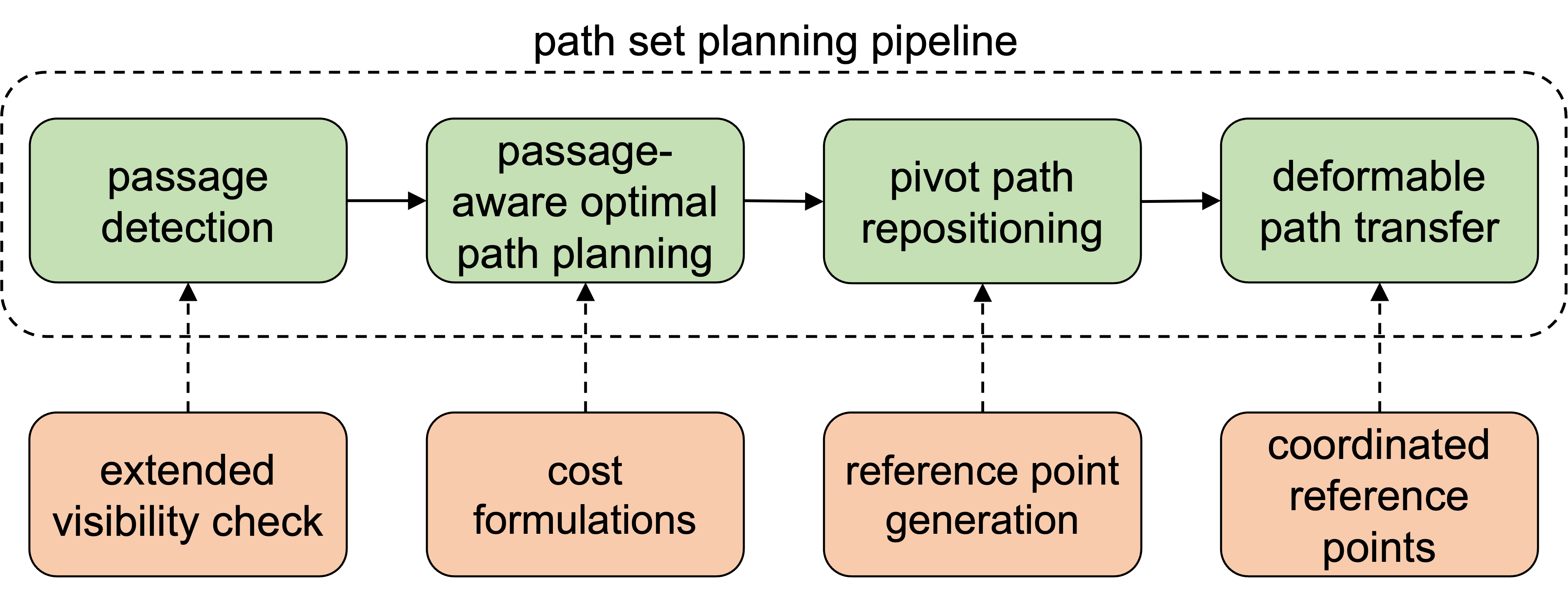}
     \endminipage \hfill
     \caption{Block diagram illustrating the overall workflow and main modules in the proposed path set planning pipeline. Blocks in the bottom row are the key components of corresponding modules.}
     \label{Path Set Planning Pipeline}
\end{figure}

In comparison to path set planning, multi-trajectory planning is more commonly studied in multi-robot systems to impose time-dependent constraints. Usually cast to spatiotemporally constrained optimization problems, multi-trajectory planning poses high complexity. In contrast, path set planning decouples time to make a more basic and tractable problem. Planned spatial paths can be converted to trajectories readily. This paper addresses path set planning, in particular a general class of homotopic path sets, and presents a systematic planning pipeline as shown in Fig. \ref{Path Set Planning Pipeline}. In obstacle-dense environments, the obstacle distribution is first perceived by identifying valid passages that constrain agent motions. A novel passage identification criterion is proposed which drastically reduces the number of passages and subsequent computations over the original visibility condition. Then, passage-aware optimal path planning is utilized to find a single path trading off classical objectives, e.g., the path length, and free space along the path to accommodate multiple paths. After analyzing the feasible topological properties of path sets, a deformable path-transfer scheme is designed to efficiently generate coordinated path sets from a single agent path. 

A recently reported study on path set planning in \cite{J. Huang 2023 TRO} only adopts DOM setups. In navigation, the concept of virtual tube in \cite{P. Mao 2022, P. Mao 2023} shows similarity to path sets in that it contains homotopic swarm trajectories, but standard planners and heavy optimization are relied on to attain trajectories. This paper extends and completes path set planning via a systematic and thorough investigation of core modules of the pipeline. The key contributions can be summarized as
\begin{enumerate}
    \item A general passage check condition to detect sparse passage distributions in environments. It reduces the passage number dramatically over pure visibility check and helps save computations in planning stages.
    \item New cost forms in passage-aware optimal path planning for adjustable planning results. Planners are enabled to optimally trade off path's accessible free space and length over conventional clearance-based methods.
    \item Further refined path transfer procedure to better guarantee transferred paths' feasibility and coordination. Path sets can be generated much more efficiently than approaches of separately planning.
\end{enumerate}
Additionally, we demonstrate extensive simulation and experimental results as well as practical robotic applications to reveal the generality and applicability of our proposals.

\section{Related Work}
The related work draws equally from manipulation planning, mostly DOM planning,  and multi-robot path planning literature. 
While DOs usually do not permit easily attainable state representations, a feasible path connecting the initial and target DO states can be planned with explicit deformation models. The DO state is usually characterized by extracted geometrical or topological properties. A survey on model-based DOM planning can be found in \cite{P. Jimenez 2012}. Models include the elasticity model \cite{F. Lamiraux 2001}, the mass-spring model \cite{R. Gayle 2005-1}, minimal-energy curves for deformable linear objects (DLOs) \cite{M. Moll 2006}, to determine valid intermediate states, namely samples. Standard planning algorithms, e.g., Probabilistic Roadmap (PRM) and Rapidly-exploring Random Tree (RRT), are utilized to attain a path to the target state. Most works are DO/task-specific, e.g., DLOs, planar DOs, and clothes  \cite{ M. Saha 2007}, \cite{A. Doumanoglou 2016}, or ad-hoc elemental action paradigms like folding and bending \cite{Liang Lu 2000}. The reliance on models or simulations severely hinders the utility of model-based planning in practice.

More relevant to this work is spatial DO path planning for a feasible path connecting initial and target DO configurations in complex environments. Using sampling-based planners, a nominal feasible path can be achieved with models/simulations \cite{O. Burchan Bayzait 2002, R. Gayle 2005}. For better practicality, DO state prediction, motion planning, and control are interleaved in \cite{D. Mcconachie 2020}. An increasing number of approaches unleash the potential of learning methods in DOM planning. In \cite{D. Mcconachie 2020 ral}, the reliability of planned state-action pairs is enforced by learning a classifier of more reliably feasible state-pairs. The imagined plan, i.e., a sequence of images to the desired goal, is attained by learning a causal InfoGAN \cite{X. Chen 2016} of the deformation dynamics and planning in the latent space in \cite{A. Wang 2019}. More recently, path sets of feedback points are leveraged as DO motion references in DOM \cite{J. Huang 2023 TRO}, but path set generation and passage processing are simplified, leading to restrictive generality.

Multi-robot path planning is a basic problem when multiple robots are present and has received substantial research in the robotics and AI community, formally named the multi-agent path finding (MAPF) problem. It aims to find feasible paths for all agents while optimizing objectives of the makespan or the sum of individual path costs \cite{A. Felner 2017, R. Stern 2019}. Numerous MAPF algorithms are proposed on common abstractions like ignoring agents' kinematics constraints and using discrete grid graphs. Despite the NP-hardness of MAPF \cite{J. Yu 2013}, sub-optimal solvers, including search-based solvers like hierarchical cooperative A$^*$ and its variants \cite{D. Silver 2005, Z. Bnaya 2014} as well as rule-based solvers \cite{M. Khorshid 2011}, can quickly find all agent paths. Some optimal solvers reduce the problem to standard and more tractable ones, e.g., integer linear programming \cite{J. Yu 2013 ICRA}, answer set programming \cite{E. Erdem 2013} and satisfiability solving \cite{P. Surynek 2012}. M$^*$ dynamically changes the dimensionality and branching factors upon detected conflicts \cite{G. Wagner 2011}. Conflict-based search adopts a two-level structure to enable fewer state examinations \cite{G. Sharon 2015}. Other methods such as increasing cost tree search \cite{G. Sharon 2013} also exist and a comprehensive survey on MAPF is available in \cite{A. Felner 2017}.

In robotics, realistic conditions in ground and aerial swarms usually render the above methods less applicable. Trajectory planning, optimization, and local motion planning are often jointly considered \cite{J. Alonso-Mora 2017}-\cite{W. Hönig 2018}. Mix-integer quadratic optimization is formulated for multiple trajectories passing specified intermediate waypoints \cite{A. Kushleyev 2013}. Discretized linear temporal logic is employed to depict robot groups' desired behaviors and trajectories are solved as a problem of satisfiability modulo theories \cite{I. Saha 2014}. An important alternative for global path planning is sampling-based methods that define the path by a set of safe team configurations. For instance, PRM is used for team formation in \cite{T. D. Barfoot 2004, A. Krontiris 2012}. Sampling-based strategies in multi-robot path planning usually need to compute cell decomposition of environments to recognize traversable areas \cite{J. Alonso-Mora 2017, N. Ayanian 2011}. Recently in \cite{P. Mao 2022, P. Mao 2023}, the virtual tube is proposed to generate infinite homotopic optimal trajectories efficiently via convex combinations of several optimal vertex trajectories. The tube's topological properties are decided by vertex paths found by RRT$^*$ (optimal RRT) that minimize the path length, but tube's accessible free space is not optimized.

\section{Passage Detection and Passage-Aware Optimal Path Planning}
This section elaborates on prerequisites for path set planning. After a brief problem statement, the extended visibility check condition in passage detection is proposed. Passage-aware optimal path planning is then presented.

\subsection{Homotopic Path Set Planning Problem}
Consider a team of $K$ agents $S = [\mathbf{s}_1\T, ..., \mathbf{s}_K\T]\T$ with $\mathbf{s}_{i} \in \mathbb{R}^3$ denoting the $i$-th agent's position. In manipulation, the team can be the group of keypoints picked on DOs for deformation description. The \textit{path set}, i.e., the collection of agent paths, is employed as the path embodiment of the team. Given the initial $S_0 = [\mathbf{s}_{0,1}\T, ..., \mathbf{s}_{0,K}\T]\T$ and target $S_d = [\mathbf{s}_{d,1}\T, ..., \mathbf{s}_{d,K}\T]\T$, the aim is to find the path set to encode the team's spatial path in complex obstacle-dense environments. 
Apart from individual path's properties, the path set needs to fulfill certain constraints as a whole. In particular, we target homotopic path sets in which all paths are homotopic. Denote $\sigma_i$ the path of $\mathbf{s}_i$, then the path set $\Sigma_S $ satisfies
\begin{equation}
\label{Problem Formulation}
\begin{split}
    \Sigma_S = \{\sigma_1, \sigma_2, ..., \sigma_K\}, \; \mathcal{H}(\sigma_i, \sigma_j) = \text{True} \; \forall \sigma_i, \sigma_j
\end{split}
\end{equation}
where $\mathcal{H}(\cdot)$ checks path homotopy. Meanwhile, $\Sigma_S$ is required to occupy a large free space in complex environments for multiple path accommodation and have short paths as detailed later. Intuitively, homotopic path sets do not allow agents to be split apart by obstacles, a widely applicable homotopy path class in DOM \cite{J. Huang 2023 TRO} and robot navigation \cite{J. Alonso-Mora 2017,  B. Zhou 2020, B. Zhou 2021}, and are also readily extendable to general cases by allowing multiple homotopic path sets \cite{P. Mao 2023, L. Jaillet 2008}.
\begin{figure}[t]
    \minipage{1 \columnwidth}
     \centering
     \includegraphics[width= 0.98 \columnwidth]{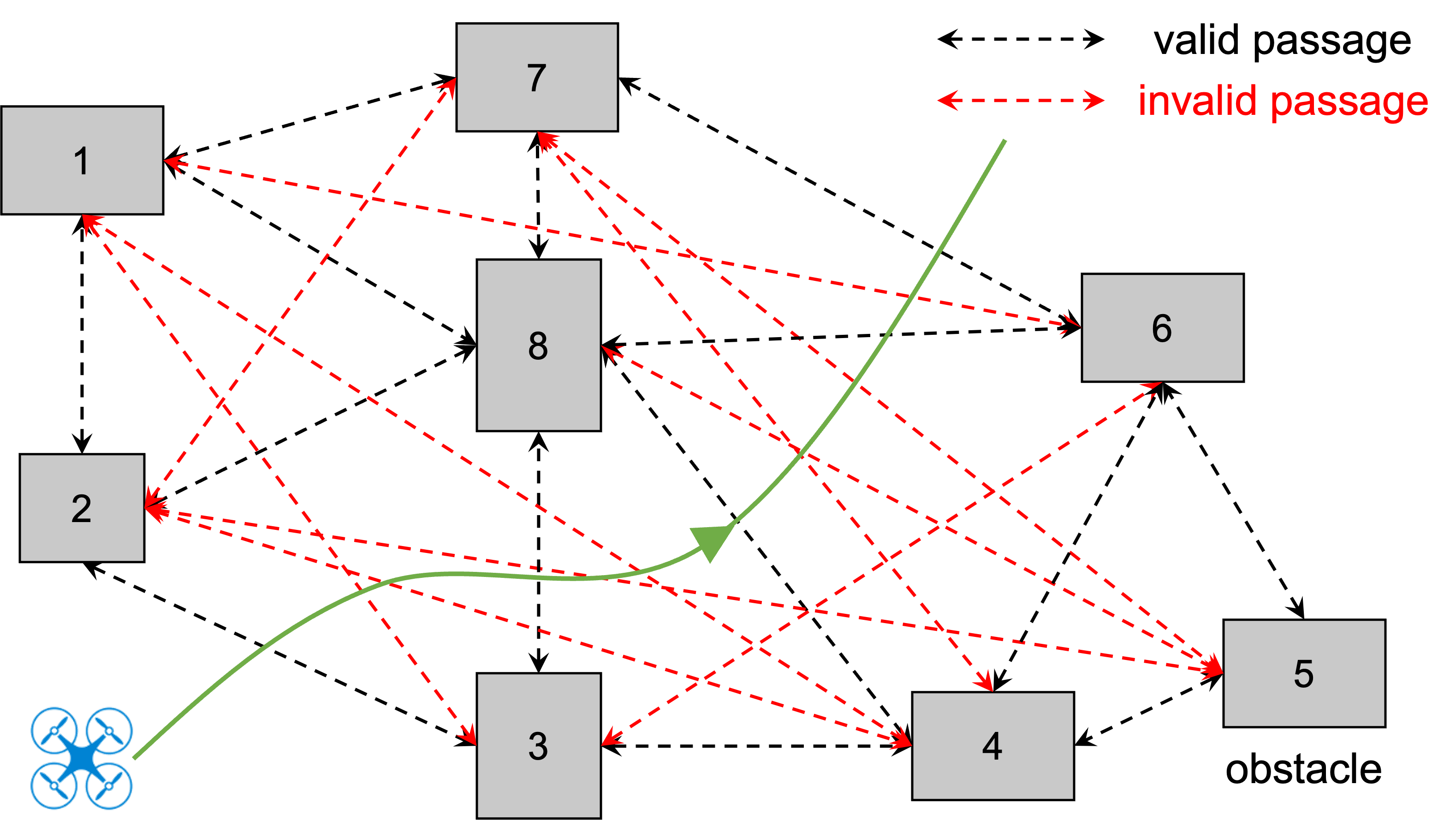}
     \endminipage \hfill
     \caption{For simplicity, segments connecting obstacle side centers represent passages formed by two obstacles here. All passages pass the visibility check, but only black ones are useful in free space determination.}
     \label{Invalid Passage Example}
\end{figure}

\subsection{Extended Visibility Check in Passage Detection}
Narrow space easily causes collisions in team motion and manipulation. Thus, sufficient free space along the path set represents a fundamental requirement.  As paths are homotopic, it suffices to resort to one agent path for a large free space. An effective way to gauge accessible free space along a path is by checking traversed passages. Unlike passage detection based on computationally intensive bridge tests \cite{Z. Sun 2005}, \cite{P. Guo 2023}, obstacle distribution is exploited for fast detection. Assume obstacles are separate polyhedrons $\mathcal{E}_i (i = 1, ..., M)$. Each ordered pair $\mathcal{E}_i, \mathcal{E}_j$ ($i < j$) forms a generic passage denoted as $(\mathcal{E}_i, \mathcal{E}_j)$. A total of $(^M_{\,2})$ such passages exist, but not all are physically valid. The pruning strategy of visibility check is used in \cite{J. Huang 2023 TRO}. $(\mathcal{E}_i, \mathcal{E}_j)$ is classified as valid only if the passage segment is collision-free with other obstacles. 
Though visually invalid passages are excluded, a significant downside is that it leaves a large fraction of passages invalid for free space determination in obstacle-dense environments, e.g., Fig. \ref{Invalid Passage Example}.

Redundant passages will incur a large computational load in following passage-related procedures in path set planning. To address this, an extended visibility check condition is proposed here to enable a thorough passage check. The shortest segment between obstacles is leveraged as a compact representation of the passage, i.e., 
\begin{equation}
\label{Nearest Point Passage Definition}
\begin{split}
    (\mathcal{E}_i, \mathcal{E}_j) &= l(\mathbf{p}_i^*, \mathbf{p}_j^*) \\
    \st  \; (\mathbf{p}_i^*, \mathbf{p}_j^*) &= \operatorname*{arg \, min}_{\mathbf{p}_i \in \mathcal{E}_i, \mathbf{p}_j \in \mathcal{E}_j} \| \mathbf{p}_i - \mathbf{p}_j \|_2
\end{split}
\end{equation}
where $l(\cdot) \subset \mathbb{R}^3$ is the segment connecting two points. In pure visibility check, $\mathcal{V}(\mathcal{E}_i, \mathcal{E}_j, \mathcal{E}_k)$ returns true if $(\mathcal{E}_i, \mathcal{E}_j)$ is not occluded by $\mathcal{E}_k$ and false otherwise
\begin{equation}
\label{Original Extended Visibility}
    \mathcal{V}(\mathcal{E}_i, \mathcal{E}_j, \mathcal{E}_k) = \text{False} \;\; \text{if} \;\; \mathcal{E}_k \cap l(\mathbf{p}_i^*, \mathbf{p}_j^*) \neq \emptyset.
\end{equation}
$(\mathcal{E}_i, \mathcal{E}_j)$ is evaluated as a passage if $\mathcal{V}(\mathcal{E}_i, \mathcal{E}_j, \mathcal{E}_k)$ is true for all $\mathcal{E}_k, k \neq i, j$.
This condition, however, is overly restrictive to filter out invalid passages. For instance, $(\mathcal{E}_2, \mathcal{E}_4)$ in Fig. \ref{Invalid Passage Example} is marked as a passage under the visibility criterion. Nonetheless, suppose an agent with a certain volume is passing $(\mathcal{E}_2, \mathcal{E}_4)$, its motion is more directly restricted by nearby obstacles $\mathcal{E}_3$ and $\mathcal{E}_8$ in passages $(\mathcal{E}_2, \mathcal{E}_3)$ and $(\mathcal{E}_3, \mathcal{E}_8)$.

\begin{figure}[t]
    \minipage{1 \columnwidth}
     \centering
     \includegraphics[width= 0.7 \columnwidth]{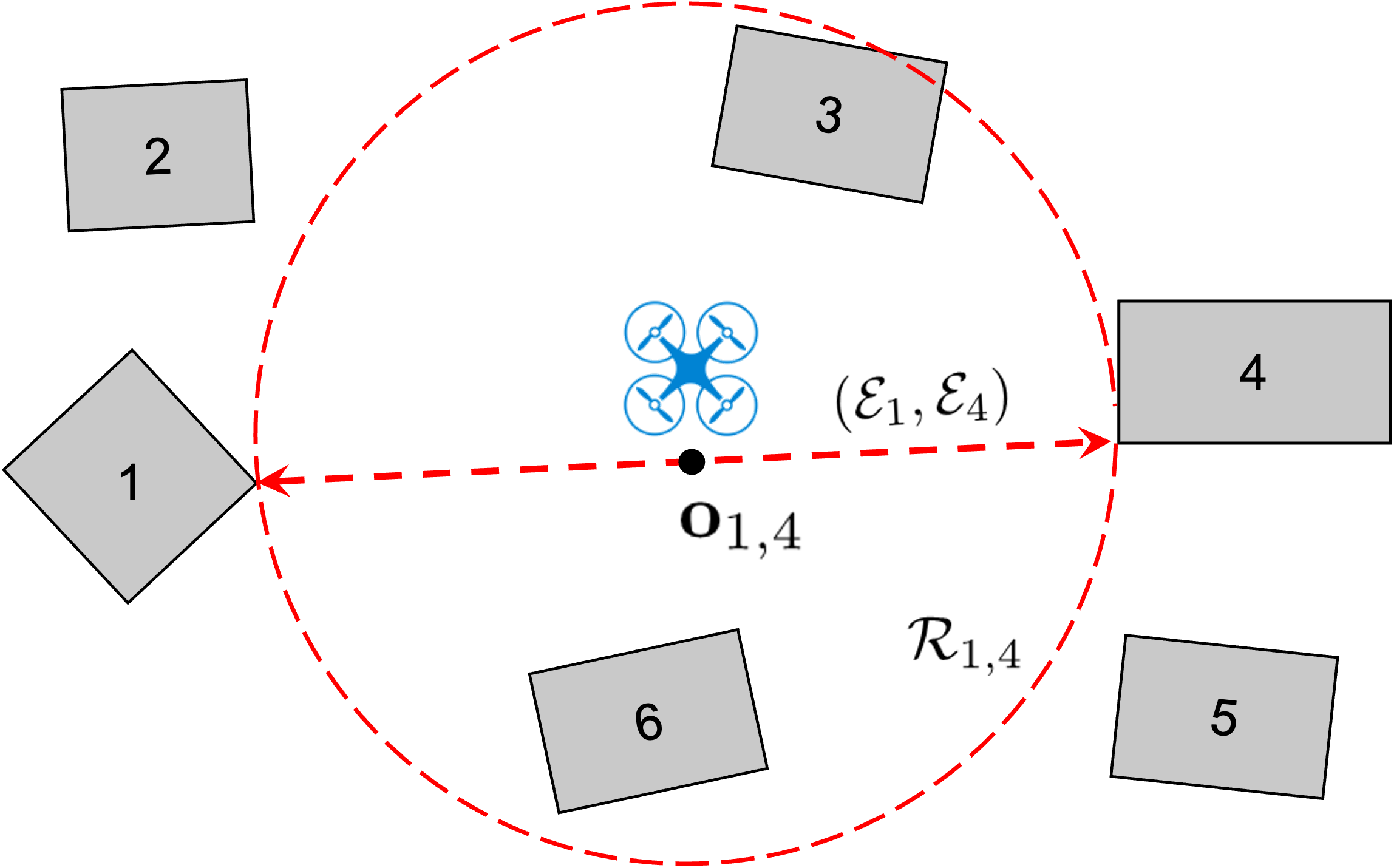}
     \endminipage \hfill
     \caption{$(\mathcal{E}_1, \mathcal{E}_4)$ will pass the original visibility check, but cannot pass the extended variant because both $\mathcal{E}_3$ and $\mathcal{E}_6$ intersect $\mathcal{R}_{1,3}$.}
     \label{Extended Visibility Condition Example}
\end{figure}

Taking the perspective of an agent with isotropic motion directions, $(\mathcal{E}_i, \mathcal{E}_j)$ constrains the agent in a circular area $\mathcal{R}_{i,j}$. The region center $\mathbf{o}_{i,j}$ coincides with the passage center and the diameter $2r_{i,j}$ equals the passage width $\| (\mathcal{E}_i, \mathcal{E}_j) \|_2$. If any other $\mathcal{E}_k$ intersects $\mathcal{R}_{i,j}$, the free space shrinks as exemplified in Fig. \ref{Extended Visibility Condition Example} for a drone. $(\mathcal{E}_i, \mathcal{E}_j)$ should be classified as invalid since the agent is more directly confined by $\mathcal{E}_k$ when it is in $(\mathcal{E}_i, \mathcal{E}_j)$. This criterion naturally involves the original visibility condition as an extended variant and can be expressed as
\begin{equation}
\label{Extended Visibility Check}
    \mathcal{V}(\mathcal{E}_i, \mathcal{E}_j, \mathcal{E}_k) = \text{False} \;\; \text{if} \;\; \mathcal{E}_k \cap \mathcal{R}_{i,j} \neq \emptyset.
\end{equation}
This infers that an agent in cluttered environments is more likely to be blocked by obstacles in its vicinity than obstacles forming the generic passage it is traversing. (\ref{Extended Visibility Check}) can also be interpreted by Voronoi diagrams that partition the environment by clearance to obstacles \cite{P. Bhattacharya 2007, H. Niu 2019, J. Hu 2020}. Specifically, (\ref{Extended Visibility Check}) imposes that $(\mathcal{E}_i, \mathcal{E}_j)$ is valid if it only goes through the Voronoi cells associated with $\mathcal{E}_i$ and $\mathcal{E}_j$. In this way, redundant passages are thoroughly excluded, which helps save computations in following passage-related procedures. In 3D maps, passage detection can be performed in height intervals determined by obstacles to maintain a sparse passage structure.

\subsection{Passage-Aware Optimal Path Planning}
Passage determination preprocesses the environment before path planning. For each $\mathbf{s}_i \in S$, the path from its start $\mathbf{s}_{0, i}$ to the target $\mathbf{s}_{d, i}$ optimizing a user-defined cost function is obtainable by optimal planers such as sampling-based methods \cite{Karaman S. 2011 ICRA}-\cite{J.D. Gammell 2014}. Formally, a feasible path is a continuous function $\sigma: [0, 1] \mapsto \mathcal{X}_{free}$ where $\mathcal{X}_{free}$ is the obstacle-free configuration space. The path argument $\tau \in [0,1]$ is given by path length parameterization by default. As aforementioned, apart from the typical criterion of path length, one core requirement for an agent path is to optimize the accessible free space for the path set. The passages passed by $\sigma$ from the start $\sigma(0)$ to $\sigma(\tau)$ are stored in an ordered list $P_{\sigma}(\tau) = \{(\mathcal{E}_i, \mathcal{E}_j), ...,  (\mathcal{E}_p, \mathcal{E}_q)\}$. $P_{\sigma}(\tau, i)$ indexes the $i$-th passage in $P_{\sigma}(\tau)$. $\min \| P_{\sigma}(\tau) \|_2$ returns the minimum passage width in $P_{\sigma}(\tau)$.

Optimal planners asymptotically find the path that optimizes some properly defined cost. A well-formulated cost is therefore essential to depict the aforementioned planning requirements. However, it is not straightforward since these requirements may be inconsistent and conflicting. A path minimizing the path length need not have sufficient free space along it and vice versa. The cost to trade off them in \cite{J. Huang 2023 TRO} is 
\begin{equation}
\label{Cost Function for Passage Passing}
    f(\sigma) = \text{Len}(\sigma) / f_P(\sigma)
\end{equation}
where Len$(\sigma)$ is the path length. $f_P(\sigma) = \min \| P_{\sigma}(1) \|_2$ is the minimum passage width passed by $\sigma$. Proportional comparison between Len$(\sigma)$ and $f_P(\sigma)$ is adopted in (\ref{Cost Function for Passage Passing}). The composite cost tends to minimize the path length while maximizing the minimum passage width the path passes.  
\begin{algorithm}[t]
    \nl Input $\mathbf{s}_{near} \in S_{near}, \mathbf{s}_{new}, P_{valid}, E$\;
    \nl \ForEach{$(\mathcal{E}_i, \mathcal{E}_j) \in P_{valid}$} {
        \nl \If {$edge(\mathbf{s}_{near}, \mathbf{s}_{new})$ \textnormal{passes} $(\mathcal{E}_i, \mathcal{E}_j)$} {
            \nl $\sigma' \leftarrow \sigma_{near}^* * edge(\mathbf{s}_{near}, \mathbf{s}_{new})$\;
            \nl Compute  $f(\sigma'), f_P(\sigma')$\;
            \If{$f(\sigma_{temp}) < f(\sigma_{new}^*)$} {
                \nl Update $f(\sigma_{new}^*), f_P(\sigma_{new}^*)$ as values of $\sigma'$\;
                \nl Update $f_{cur}(\mathbf{s}_{new})$ and parent  of $\mathbf{s}_{new}$\;
                \nl $E \leftarrow (E \, \backslash \, \{edge(\mathbf{s}_{parent}, \mathbf{s}_{new})\}) \, \cup \, \{edge(\mathbf{s}_{near}, \mathbf{s}_{new})\}$\;
            }
        }
    }
    \caption{Update Cost of $\mathbf{s}_{new}$ in A New Edge}
    \label{Update Node Cost}
\end{algorithm}

Terms' priorities in (\ref{Cost Function for Passage Passing}) are fixed. Adjustable costs are more desirable to enable different path preferences considering the free space requirement varies with problem setups such as the team size and the obstacle density. A weighted cost structure is introduced herein as
\begin{equation}
\label{Weighted Cost Function for Passage Passing}
    f(\sigma) = \text{Len}(\sigma) - k_P f_P(\sigma)
\end{equation}
where $k_P > 0$ acts as the weight of $f_P(\sigma)$ that determines the importance of $f(\sigma)$. Intuitively, the passage width is converted into a generalized and weighted path length subtracted from the true path length in $(\ref{Weighted Cost Function for Passage Passing})$. By taking different $k_P$, the dominance between Len$(\sigma)$ and $f_P$ changes. $k_P$ selection is problem-specific. In most scenarios, $\| P_\sigma(1) \|_2$ is significantly smaller than Len$(\sigma)$, $k_P$ should not be too small to bring out the effect of $f_P(\sigma)$ in the cost.

The cost formulations (\ref{Weighted Cost Function for Passage Passing}) are monotonic under path concatenation (Len$(\cdot)$ is monotonically increasing, $f_P(\cdot)$ is monotonically non-increasing) and bounded. Therefore, sampling-based optimal planners are guaranteed to find the optimal path asymptotically \cite{Karaman S. 2011} and RRT$^*$ is taken in our implementation. Denote $\sigma_{new}^*$ the optimal path from the start $\mathbf{s}_{init}$ to the new sample $\mathbf{s}_{new}$. $\mathbf{s}_{new}$ carries attributes of $f(\sigma_{new}^*)$, $f_P(\sigma_{new}^*)$ and $f_{cur}(\mathbf{s}_{new})$, the passage width passed by the edge from the parent node to $\mathbf{s}_{new}$. $f_P(\sigma_{new}^*)$ and $f_{cur}(\mathbf{s}_{new})$ are initialized as large values to indicate no passing of passages. To iteratively find $\sigma_{new}^*$, passage passing is checked in every attempt to add the edge between $\mathbf{s}_{new}$ and a near node $\mathbf{s}_{near}$. Attributes are updated accordingly (see Algorithm \ref{Update Node Cost}). This procedure is invoked when finding the parent node and rewiring the tree, i.e., \texttt{GetParent()} and \texttt{Rewire()}. See Algorithm \ref{RRTStar} for passage-aware optimal path planning for a single path that integrates the extended visibility check for passage detection and the new cost formulation. 
\begin{algorithm}[t]
\nl $P_{valid} \leftarrow \texttt{ExtendedVisibilityCheck}(\mathcal{E}_1, ..., \mathcal{E}_M)$\;
\nl $V \leftarrow \{ \mathbf{s}_{init} \}$; $E \leftarrow \emptyset$\;
\nl \For {$i = 1, 2, ..., N$} {
    \nl $\mathbf{s}_{rand} \leftarrow \texttt{SampleFree}(\delta)$\;
    \nl $\mathbf{s}_{nearest} \leftarrow \texttt{Nearest}(G = (V, E), \mathbf{s}_{rand})$\;
    \nl $\mathbf{s}_{new} \leftarrow \texttt{Steer}(\mathbf{s}_{nearest}, \mathbf{s}_{rand})$\;
    \nl \If{\textnormal{\texttt{ObstacleFree}$(\mathbf{s}_{nearest}, \mathbf{s}_{new})$}} {
        \nl $S_{near} \leftarrow \texttt{Near}(G = (V, E), \mathbf{s}_{new}, r_{near})$\;
        \nl $V \leftarrow V \cup \{ \mathbf{s}_{new} \}$\;
        \nl $\mathbf{s}_{min} \leftarrow \texttt{GetParent}(S_{near}, \mathbf{s}_{new}, P_{valid})$\;
        \nl $E \leftarrow E \cup \{edge(\mathbf{s}_{min}, \mathbf{s}_{new})\}$\;
        \nl $\texttt{Rewire}(G = (V, E), S_{near}, \mathbf{s}_{min}, \mathbf{s}_{new}, P_{valid})$\;
    }
    }
    \nl \Return $G = (V, E)$\;
    \caption{RRT$^*$-Based Passage-Aware Optimal Path Planning}
    \label{RRTStar}
\end{algorithm}

\section{Path Set Homotopy and Path Transfer}
This section first discusses the feasibility requirements for path sets regarding paths' homotopic properties. Path transfer is then introduced as the basic primitive for path set generation.

\subsection{Feasibility Requirement for Path Sets in Homotopy}
For the feasibility of the path set $\Sigma_S$ as a collection, paths' homotopic interrelationships are analyzed. In general non-winding scenarios where agents do not wrap around obstacles, the homotopy constraint is imposed on paths. $\sigma_1, \sigma_2$ with identical initial and final positions are path homotopic if there exists a continuous function, i.e., the homotopy, $\psi(\cdot): [0, 1] \mapsto \Sigma_{free}$, where $\Sigma_{free}$ is the set of paths in $\mathcal{X}_{free}$, such that $\psi(0) = \sigma_1, \psi(1) = \sigma_2$ and $\psi(x) \in \Sigma_{free}, \, \forall \, x \in [0,1]$. Homotopic paths essentially can continuously deform to one another in $\mathcal{X}_{free}$ but are not easily verifiable in general. For easy verifiability and good generality, straight-line homotopy is imposed on paths. If  $\sigma_1, \sigma_2$ are path homotopic, their straight-line homotopy is
\begin{equation}
\label{Straight-line Homotopy}
    \psi_{1,2}(x, \tau) = (1 - x) \sigma_1(\tau) + x \sigma_2(\tau), \;\; x, \tau \in [0, 1].
\end{equation}
$\sigma_1, \sigma_2$ are said to be strong path homotopic if $\psi_{1,2}(x, \tau) \in \mathcal{X}_{free}$ always holds, indicating that the hypersurface swept by $\psi_{1,2}(x, \tau)$ lies in $\mathcal{X}_{free}$. We do not construct strictly homotopic paths with the same endpoints as in \cite{J. M. Esposito 2006}.
$\sigma_i, \sigma_j \in \Sigma_S$ are said to be strong path homotopic-like if their straight-line homotopy in (\ref{Straight-line Homotopy}) remains in $\Sigma_{free}$.

\begin{figure}[t]
    \minipage{1 \columnwidth}
    \centering
    \includegraphics[width= 0.83 \columnwidth]{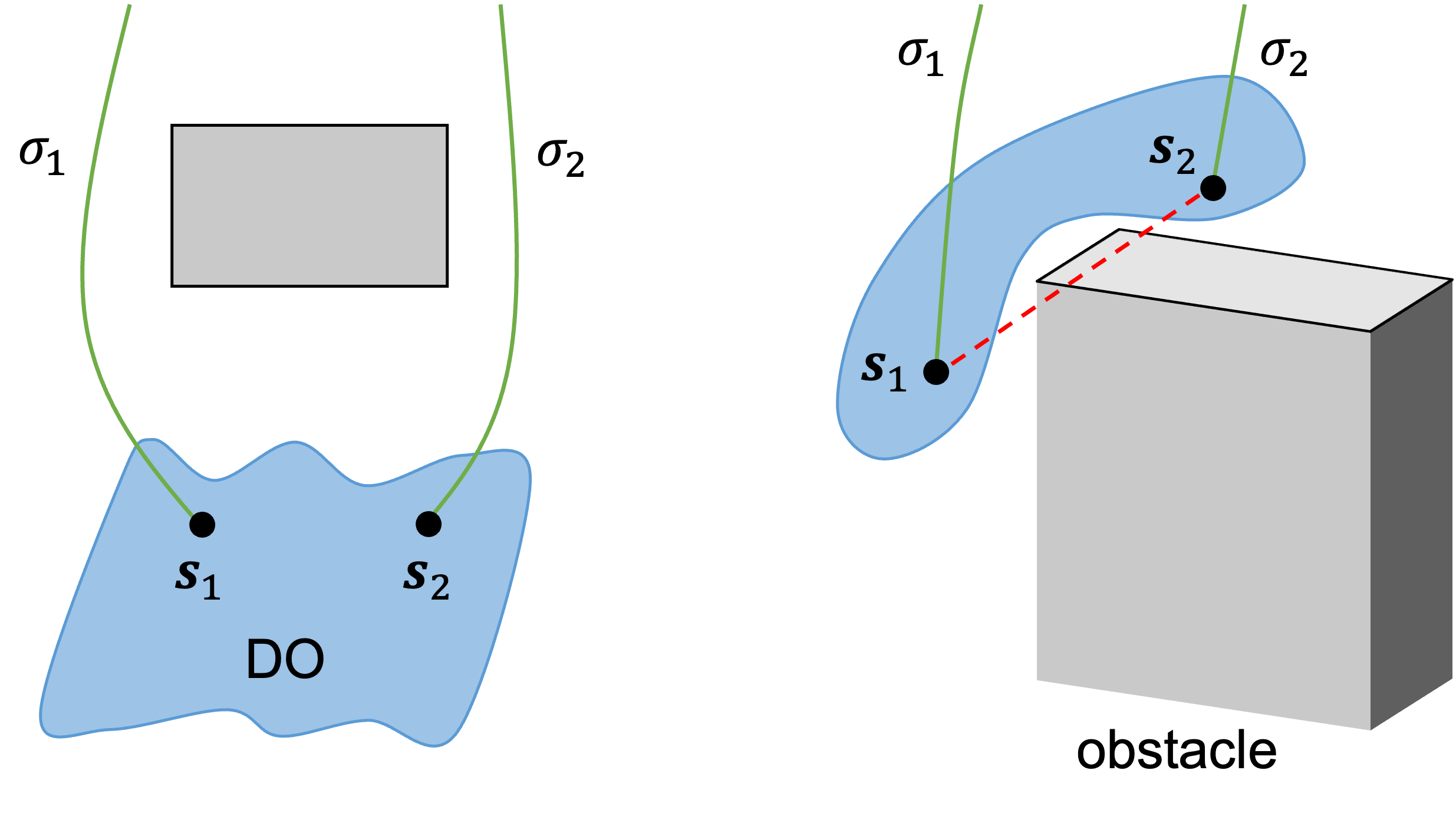}    
    \endminipage \hfill
    \caption{In the left 2D case, DO point path 1 and 2 are not homotopic and thus infeasible. In the right 3D case, though path 1 and 2 are not strong path homotopic-like, the shown DO pose is feasible.}
    \label{Path Homotopy}
\end{figure}

Any pair of paths in $\Sigma_S$ are required to be strong path homotopic-like as in \cite{J. Huang 2023 TRO}, equivalent to the uniform visibility condition for quadrotor paths \cite{B. Zhou 2020, B. Zhou 2021}, line of sight \cite{J. M. Esposito 2006}, and visibility deformation of roadmaps \cite{L. Jaillet 2008}. Since agents are not separated by obstacles, paths are homotopic to transform into each other, which is hard to check in high-dimensional spaces. The strong homotopic-like condition is easy to check, although it can be overly constrained in 3D space to exclude feasible situations where paths round obstacles like Fig. \ref{Path Homotopy}. This can be further checked by examining if the straight-line homotopy passes the entire obstacle top part.

\subsection{Path Transfer}
Generating $\Sigma_S$ by planning each agent path in a decentralized fashion is complex due to the homotopy constraint. Coordination among paths is also hard to achieve. 
Paths in a homotopic path set share an identical passage passing list in 2D, i.e., $P_{\sigma_i} = P_{\sigma_j}, \forall \, \sigma_i, \sigma_j \in \Sigma_S$, or have limited differences in 3D. Therefore, if one path in $\Sigma_S$ is planned, other paths can be generated by transferring from it. This flow fulfills homotopy by construction and is computationally efficient. $S_0$ and $S_d$ are essentially two clusters with different agent distributions. A path set from $S_0$ to $S_d$ connects the two clusters while meeting the above constraints.  Suppose $\mathbf{s}_p \in S$ has a planned path $\sigma_p$ from $\mathbf{s}_{0,p}$ to $\mathbf{s}_{d,p}$, then for each point $\mathbf{s}_i \in S$, the path transferred from $\sigma_p$ is
\begin{equation}
\label{General Path Trasfer}
    \sigma_{t,i}(\tau_i) = \sigma_p(\tau_p) + \mathbf{v}_{p \rightarrow i}(\tau_p, \tau_i)
\end{equation}
where a general path transfer form is utilized compared to \cite{J. Huang 2023 TRO}. It permits different path arguments in $\sigma_p$, $\sigma_{t, i}$, and a varying transfer vector to enable more flexible transfer.

The planned path $\sigma_p$ in path transfer is conceptually similar to the generator curve in \cite{P. Mao 2023}. The corresponding agent, termed \textit{pivot}, can be picked arbitrarily since following transfer procedures are invariant to pivot. Designate $p$ the chosen pivot index. The pivot path $\sigma_p$ is found by the passage-aware optimal path planner. When transferring $\sigma_p$ to other agents, forward and backward transfer are proposed in \cite{J. Huang 2023 TRO} by assigning $\mathbf{v}_{p \rightarrow i}$ as $\mathbf{v}_{0, p \rightarrow i} = \mathbf{s}_{0, i} - \mathbf{s}_{0, p}$ and $\mathbf{v}_{d, p \rightarrow i} = \mathbf{s}_{d, i} - \mathbf{s}_{d, p}$, respectively, where extra postprocessing of path concatenation is required. To resolve this, these two transfer paradigms are combined in (\ref{General Path Trasfer}) to be 
\begin{equation}
\label{Combined Path Trasfer}
    \sigma_{t, i}(\tau) = \sigma_p(\tau) + (1 - \tau) \mathbf{v}_{0, p \rightarrow i} + \tau \mathbf{v}_{d, p \rightarrow i}
\end{equation}
where $\tau_i = \tau_p = \tau$, $\mathbf{v}_{0, p \rightarrow i}$ and $\mathbf{v}_{d, p \rightarrow i}$ are linearly interpolated to compose a varying transfer vector along $\sigma_{t, i}$. In this way, $\sigma_{t, i}(0) = \mathbf{s}_{0,i}$, $\sigma_{t, i}(1) = \mathbf{s}_{d,i}$ with no need for path postprocessing. Performing $(\ref{Combined Path Trasfer})$ for each agent in $S$ leads to a transferred path set $\Sigma_t(S_0, S_d, \sigma_p)$. $\Sigma_t$ is strong homotopic-like if the hyperplane swept by the varying transfer vector $\mathbf{v}_{p \rightarrow i}$ keeps collision-free. As constrained environments are present, such an assumption usually fails, which entails refined processing to reach the final path set.

\section{Path Set Generation Using Path Transfer}
\label{Path Set Generation Section}
In this section, a path set generation scheme for constrained environments is proposed incorporating two refined steps on \cite{J. Huang 2023 TRO}: 1) pivot path planning and repositioning, and 2) coordinated deformable path transfer.

\subsection{Pivot Path Planning and Repositioning}
\subsubsection{Repositioning Reference Points Determination} 
Before planning the pivot path $\sigma_p$, a pivot selection criterion minimizing the transfer vector magnitude is introduced as
\begin{equation}
\label{Pivot Selection}
    p = \operatorname*{arg \, min}_{1 \leq i \leq K} \; \operatorname*{max}_{1 \leq j \leq K} \; ( \| \mathbf{s}_{0, i} - \mathbf{s}_{0, j} \|_2, \| \mathbf{s}_{d, i} - \mathbf{s}_{d, j} \|_2)
\end{equation}
which limits path transfer to a tunnel centered at $\sigma_p$ with a radius as small as possible. In $\Sigma_t$, transferred paths can percolate obstacles easily since $\sigma_p$ is often close to obstacles to reduce its length in (\ref{Cost Function for Passage Passing}) or (\ref{Weighted Cost Function for Passage Passing}). $\sigma_p$ thus needs to be repositioned to a more reasonable configuration while preserving $P_{\sigma_p}(1)$. After planning $\sigma_p$, $\Sigma_t$ is obtained for checking passage intersections. Usually, it is sufficient to consider obstacles in $P_{\sigma_p}(1)$, but this may miss some nearby obstacles. For completeness, a distance filter is first applied to register obstacles near $\sigma_p$. Denote the distance between $\sigma_p$ and $\mathcal{E}_i$ as 
\begin{equation}
\label{Path Obstacle Distance}
    d(\sigma_p, \mathcal{E}_i) = \min_{\tau \in [0, 1], \mathbf{p}_i \in \mathcal{E}_i} \| \sigma_p(\tau)  - \mathbf{p}_i \|_2.
 \end{equation}
A threshold $\lambda = \max_{ 1 \leq i \leq K} \max(\| \mathbf{s}_{0, i} - \mathbf{s}_{0, p} \|_2, \| \mathbf{s}_{d, i} - \mathbf{s}_{d, p} \|_2)$ is set to rule out obstacles not in $\sigma_p$'s vicinity. The remaining obstacles are divided into two categories: obstacles in passages traversed by $\sigma_p$, termed passage obstacles, and isolated obstacles otherwise. Formally, define $\mathcal{E}_{near} = \{ \mathcal{E}_i \, | \,d(\sigma_p, \mathcal{E}_i) \leq \lambda \}$, $\mathcal{E}_{P} = \{ \mathcal{E}_i \, | \, (\mathcal{E}_i, \mathcal{E}_j) \in P_{\sigma_p}(1) \, \text{or} \, (\mathcal{E}_j, \mathcal{E}_i) \in P_{\sigma_p}(1) \, \exists \, \mathcal{E}_j \}$. Passage obstacles are $\mathcal{E}_{P}^{near} = \mathcal{E}_{near} \cap  \mathcal{E}_{P}$. Isolated obstacles are $\mathcal{E}_{iso} = \mathcal{E}_{near} \setminus \mathcal{E}_{P}^{near}$.

\begin{figure}[t]
    \minipage{1 \columnwidth}
    \centering
    \includegraphics[width= 0.86 \columnwidth]{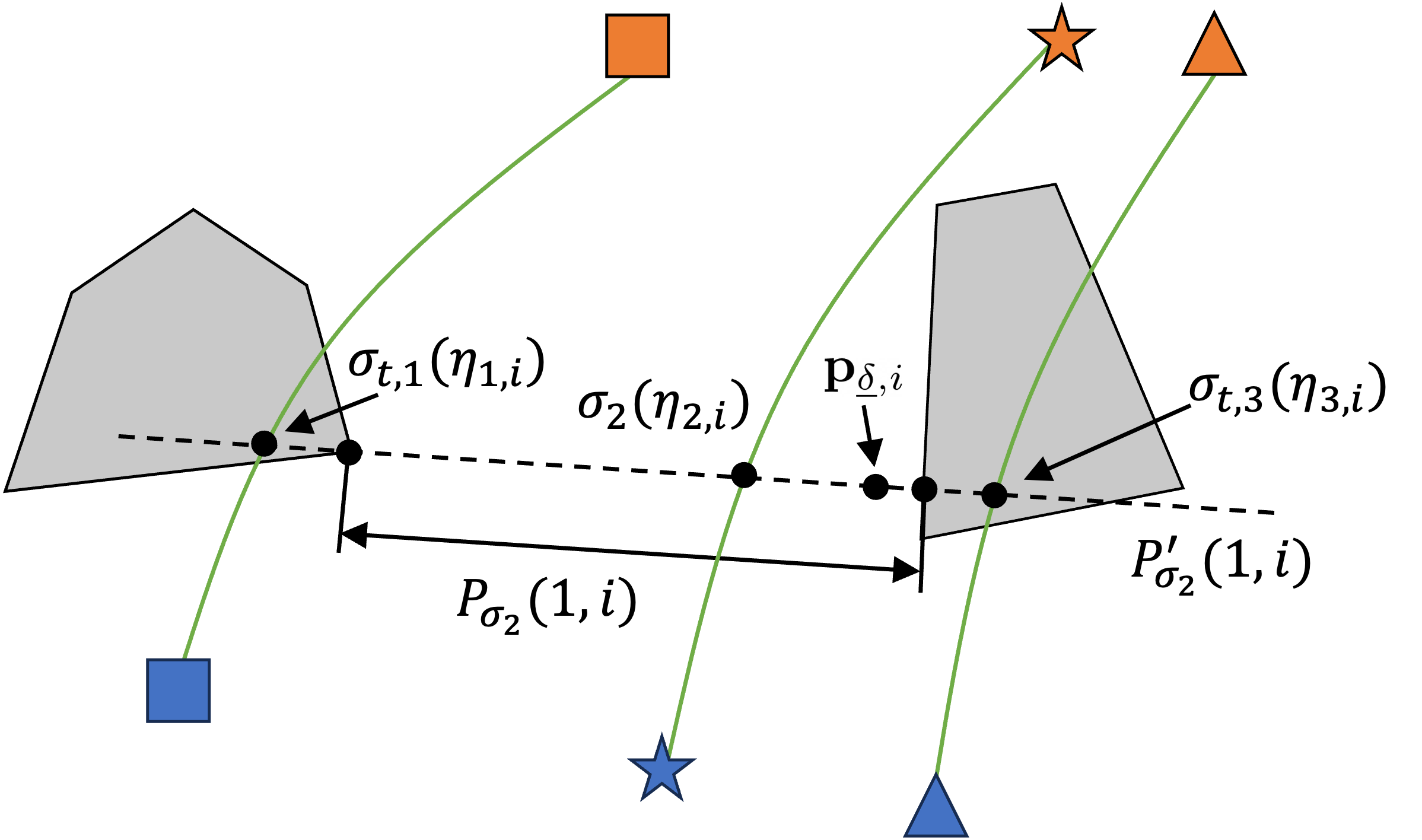}    
    \endminipage \hfill
    \caption{In the shown passage, $\sigma_2$ is the pivot path. $\sigma_{t, 1}, \sigma_{t, 3}$ are transferred from $\sigma_2$ and both collide with obstacles. To tackle this, $\sigma_2$ is repositioned first. $\sigma_{t, 1}, \sigma_{t, 3}$ are then locally deformed.}
    \label{Repositioning Scenario}
\end{figure}

\begin{figure}[t]
    \minipage{1 \columnwidth}
    \centering
    \includegraphics[width= 1 \columnwidth]{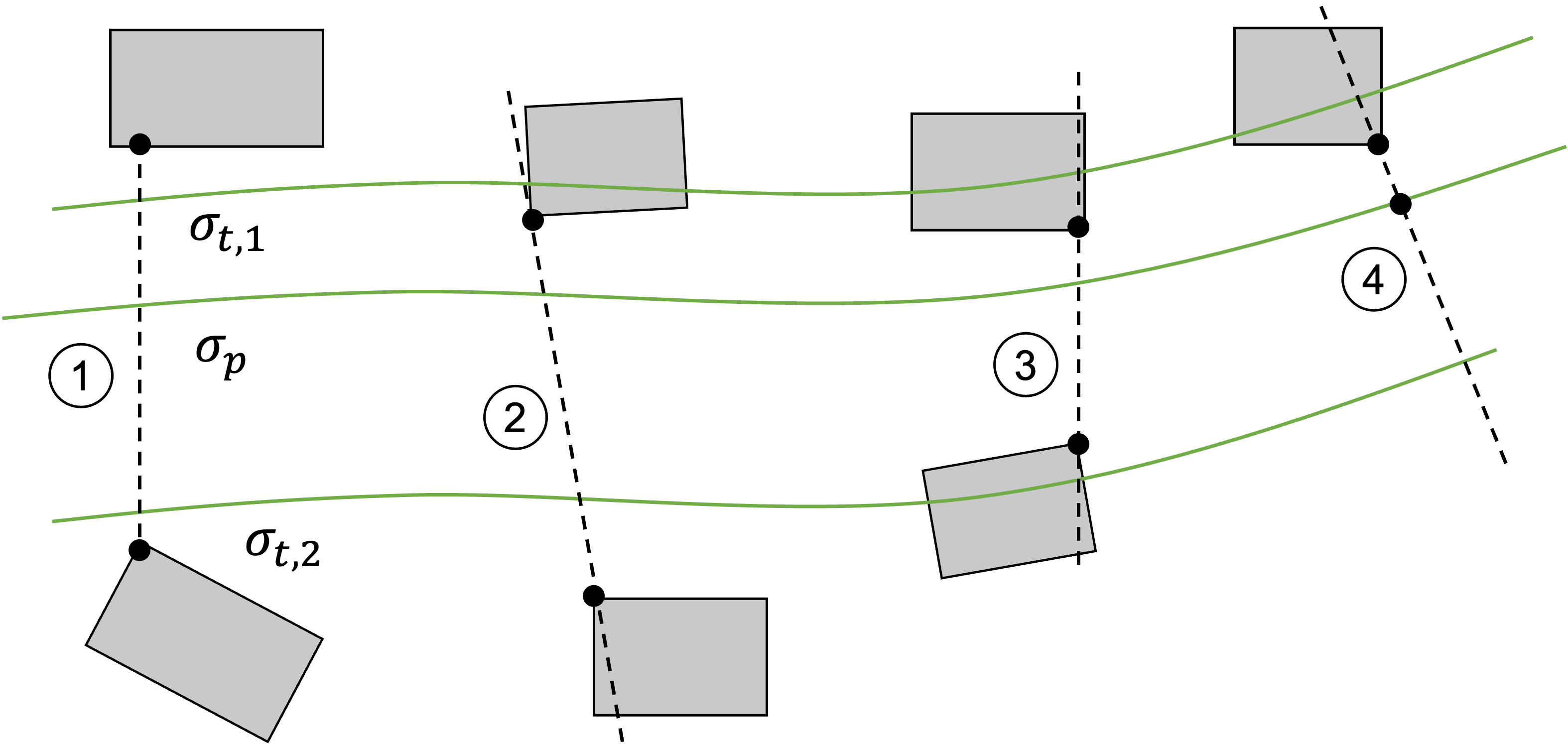}
    \endminipage \hfill
    \caption{Different situations when attaining reference points for the pivot path $\sigma_p$. 1. No need for repositioning. 2. Translate $\sigma_p$ along the passage segment. 3. Reposition $\sigma_p$ to proportionally compress the chord. 4. Push $\sigma_p$ away from an isolated obstacle.}
    \label{Transferred Path Repositioning}
\end{figure}

The relative position of $\sigma_p$ to nearby obstacles determines how $\sigma_p$ should be locally repositioned. The passage list $P_{\sigma_p}(1)$ is updated to only contain passages with obstacles in $\mathcal{E}_{near}$. $P_{\sigma_p}(1, i) = (\mathcal{E}_j, \mathcal{E}_k)$ is discarded if both $\mathcal{E}_j$ and $\mathcal{E}_k$ are not in $\mathcal{E}_{near}$. The intersection segment between $\Sigma_t$ and $P_{\sigma_p}(1, i)$ is characterized by the chord, denoted as $\Sigma_t \cap P_{\sigma_p}'(1, i)$, with its length being
\begin{equation}
\label{Chord Length}
    \| \Sigma_t \cap P_{\sigma_p}'(1,i) \|_2 = \max_{1 \leq k, j \leq K} \| \sigma_{t, k}(\eta_{k, i}) - \sigma_{t, j}(\eta_{j, i}) \|_2
\end{equation}
where $P_{\sigma_p}'(1, i)$ signifies the entire straight line on which $P_{\sigma_p}(1, i)$ lies. $\sigma_{t, k}(\eta_{k, i})$ is the intersection point between $\sigma_{t, k}$ and $P_{\sigma_p}'(1, i)$. The ordered intersection points between $\sigma_p$ and $P_{\sigma_p}(1)$, $\{\sigma_p(\eta_{p, 1}), \sigma_p(\eta_{p, 2}), ..., \sigma_p(\eta_{p, n})\}$, constitute reference points of $\sigma_p$. 
Two overlapping possibilities between $\Sigma_t \cap P_{\sigma_p}'(1, i)$ and $P_{\sigma_p}(1, i)$ exist: $\Sigma_t \cap P_{\sigma_p}'(1, i)$ is completely contained in $P_{\sigma_p}(1, i)$ or otherwise. If $\Sigma_t \cap P_{\sigma_p}'(1, i)$ falls inside $P_{\sigma_p}(1, i)$, the intersection point $\sigma_p(\eta_{p, i})$ just remains unchanged for anchoring. In situations otherwise, the chord is in collision and $\sigma_p$ should be modified locally. If $\| \Sigma_t \cap P_{\sigma_p}'(1, i) \|_2 \leq \| P_{\sigma_p}(1, i) \|_2$, $P_{\sigma}(1, i)$ is sufficiently wide. $\sigma_p$ can be simply translated along $P_{\sigma_p}(1, i)$ to move the chord into $P_{\sigma_p}(1, i)$. Denote $\sigma_p^*(\eta_{p, i})$ the adjusted reference point, it can be given as
\begin{equation}
\label{Reference Point Position 1}
   \sigma_p^*(\eta_{p, i}) = \sigma_p(\eta_{p, i}) + \mathbf{d}_{\underline{\delta}, i} 
\end{equation}
where $\mathbf{d}_{\underline{\delta}, i}$ is the translation along $P_{\sigma_p}(1, i)$. Note that there is one chord end outside $P_{\sigma_p}(1, i)$, be it $\sigma_q(\eta_{q, i})$. After translation, the end is moved to a point $\mathbf{p}_{\underline{\delta}, i}$ on $P_{\sigma_p}(1, i)$ with a preset clearance $\underline{\delta}$ to obstacles. The shift is $\mathbf{d}_{\underline{\delta}, i} = \mathbf{p}_{\underline{\delta}, i} - \sigma_q(\eta_{q, i})$ as illustrated in Fig. \ref{Repositioning Scenario} and Fig. \ref{Transferred Path Repositioning}. 

$\| \Sigma_t \cap P_{\sigma_p}'(1, i) \|_2 > \| P_{\sigma_p}(1, i) \|_2$ is more constrained. The chord cannot be placed within the passage segment by translation. Repositioning $\sigma_p$ now aims for a coordinated distribution of all transferred paths. The strategy to overlap centers of $\Sigma_t \cap P_{\sigma_p}'(1, i)$ and $P_{\sigma_p}(1, i)$ is utilized in \cite{J. Huang 2023 TRO}. Its problem is that it may cause extremely unbalanced path distributions. To avoid this, proportionally placing intersection points on $P_{\sigma_p}(1, i)$ according to the relative distribution of $\{\sigma_{t,1}(\eta_{1, i}), \sigma_{t, 2}(\eta_{2, i}), ..., \sigma_{t, K}(\eta_{K, i})\}$ provides more reasonable reference points. $\sigma_p^*(\eta_{p, i})$ now is
\begin{equation}
\label{Reference Point Position 2}
   \sigma_p^*(\eta_{p, i}) = \mathbf{p}_{\underline{\delta}, i} + r_i (\sigma_p(\eta_{p, i}) - \sigma_q(\eta_{q, i}))
\end{equation}
where $r_i = ({\| P_{\sigma_p}(1, i) \|_2 - 2 \underline{\delta}}) / {\| \Sigma_t \cap P_{\sigma_p}'(1, i) \|_2}$ is the scaling ratio. $\mathbf{p}_{\underline{\delta}, i}$ is the new chord end. As $\| \Sigma_t \cap P_{\sigma_p}'(1, i) \|_2 > \| P_{\sigma_p}(1, i) \|_2$, at least one chord end is outside of $P_{\sigma_p}(1, i)$ to be taken as $\sigma_q(\eta_{q, i})$. For isolated obstacles, we only consider the case in which they collide with some path or lie between paths. As such, $\sigma_p(\tau_{\mathcal{E}_i})$ is translated along the direction of $\sigma_p(\tau_{\mathcal{E}_i}) - \mathbf{p}_i^*$, where $\sigma_p(\tau_{\mathcal{E}_i})$ is the minimum distance projection of $\mathcal{E}_i$ on $\sigma_p$ in (\ref{Path Obstacle Distance}). $\mathbf{p}_i^*$ is the optimal point on $\mathcal{E}_i$. This is analogous to (\ref{Reference Point Position 1}) by replacing $P_{\sigma_p}'(1, i)$ with the line of $\sigma_p(\tau_{\mathcal{E}_i}) - \mathbf{p}_i^*$.

\subsubsection{Repositioning Pivot Path}
A series of reference points have been obtained along $\sigma_p$. $\sigma_p$ is repositioned iteratively between every two consecutive reference points in a linear interpolation manner. For $\eta_{p, i} \leq \tau \leq \eta_{p, i+1}$, the new path position $\sigma_p^*(\tau)$ is given as 
\begin{multline}
\label{Reposition Pivot Path}
    \sigma_p^*(\tau) = \sigma_p(\tau) + \frac{\tau - \eta_{p, i}}{\eta_{p, i+1} - \eta_{p, i}}(\sigma_p^*(\eta_{p,i+1}) - \sigma_p(\eta_{p,i+1})) \\ + \frac{\eta_{p, i+1}  - \tau}{\eta_{p, i+1} - \eta_{p, i}}(\sigma_p^*(\eta_{p,i}) - \sigma_p(\eta_{p,i})).
\end{multline}
$\sigma_p^*$ is the repositioned $\sigma_p$. The start and final points of $\sigma_p$ need to be inserted to establish an augmented list of reference points $\{ \sigma_p(0), \sigma_p^*(\eta_{p, 1}), ..., \sigma_p^*(\eta_{p, n}), \sigma_p(1) \}$. Since the shift magnitude on a passage segment is small compared to the total path length, the path segment in (\ref{Reposition Pivot Path}) is feasible in general. As for the repositioning procedure in 3D space, the chord evolves into the convex hull formed by intersection points with the passage plane. Without considering floating obstacles, the shift of $\sigma_p(\eta_{p, i})$ is restricted in the direction parallel to the ground and rules are set similarly to (\ref{Reference Point Position 1}) and (\ref{Reference Point Position 2}).
One inherent problem of chords is that they may poorly represent how the entire path set passes through the passage and lead to nonsmooth path segments, e.g., when the passage is nearly parallel to the local path. Inspired by the virtual tube \cite{{P. Mao 2022, P. Mao 2023}}, we propose a geometrical approach for chord determination, which uses the normal direction of the pivot path to get the chord and then rotates it back to the passage segment. Details are given in Appendix \ref{App A}.
\begin{figure*}[t]
    \minipage{2 \columnwidth}
    \centering
    \subfigure[] {
    \includegraphics[width= 0.322 \columnwidth]{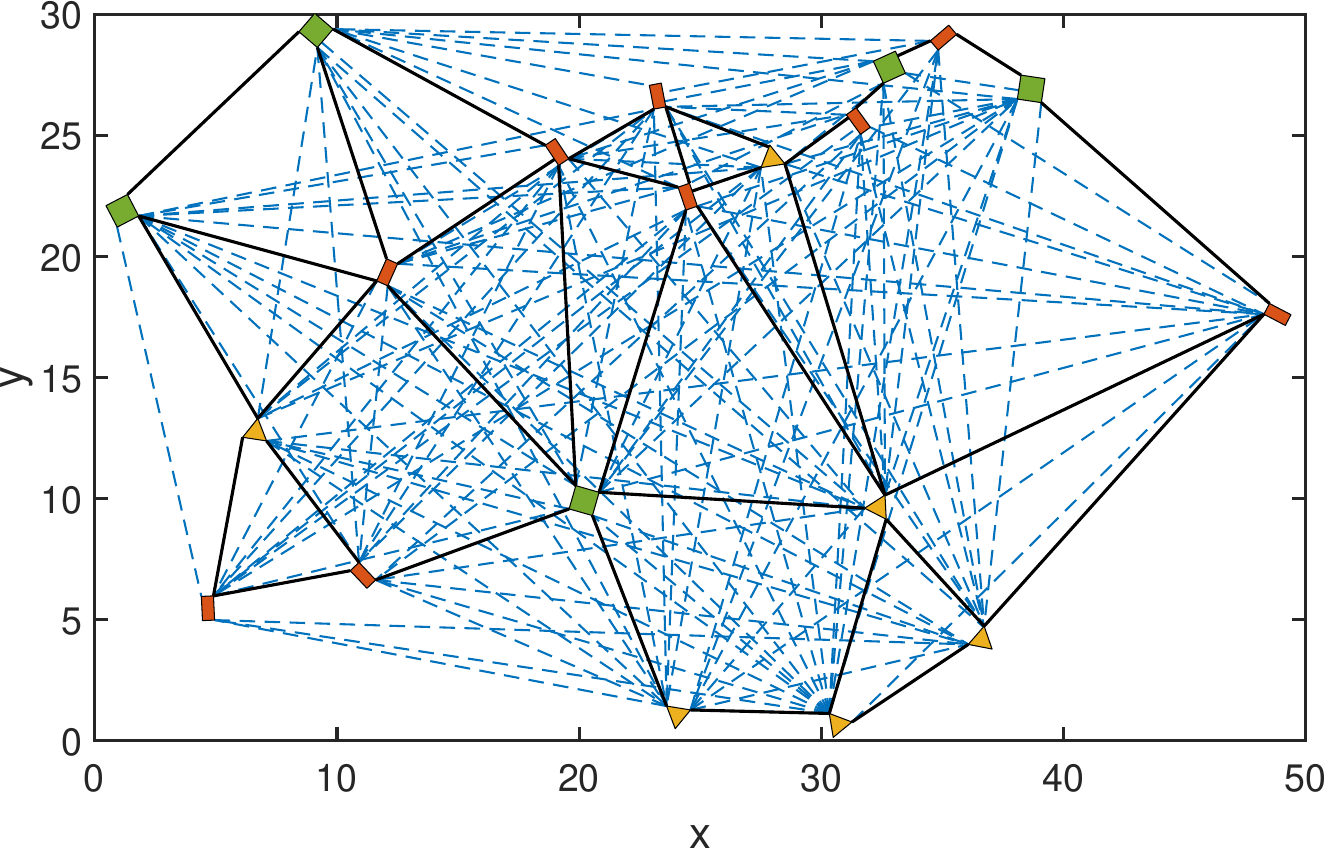}}
    \subfigure[] {
    \includegraphics[width= 0.322 \columnwidth]{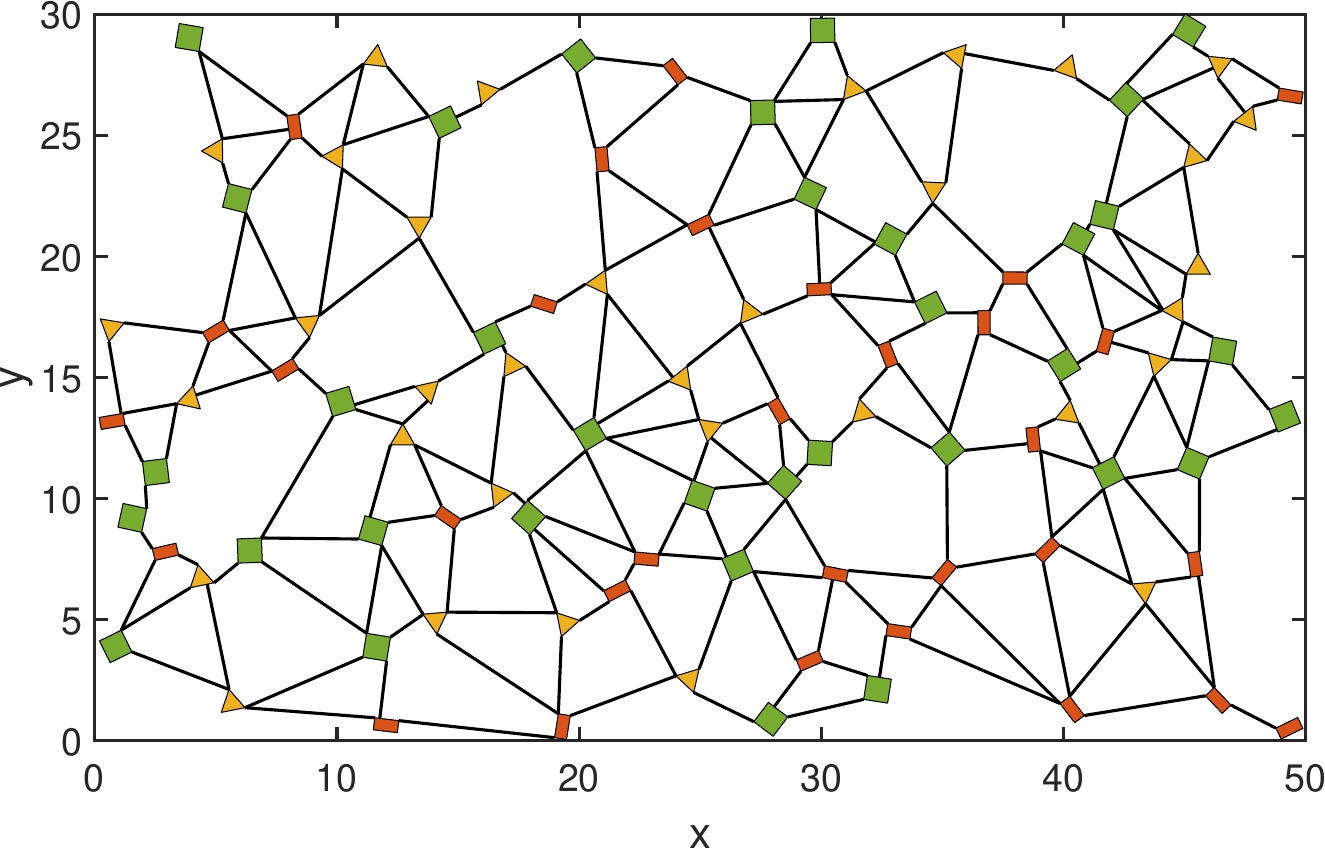}}
    \centering
    \subfigure[] {
    \includegraphics[width= 0.322 \columnwidth]{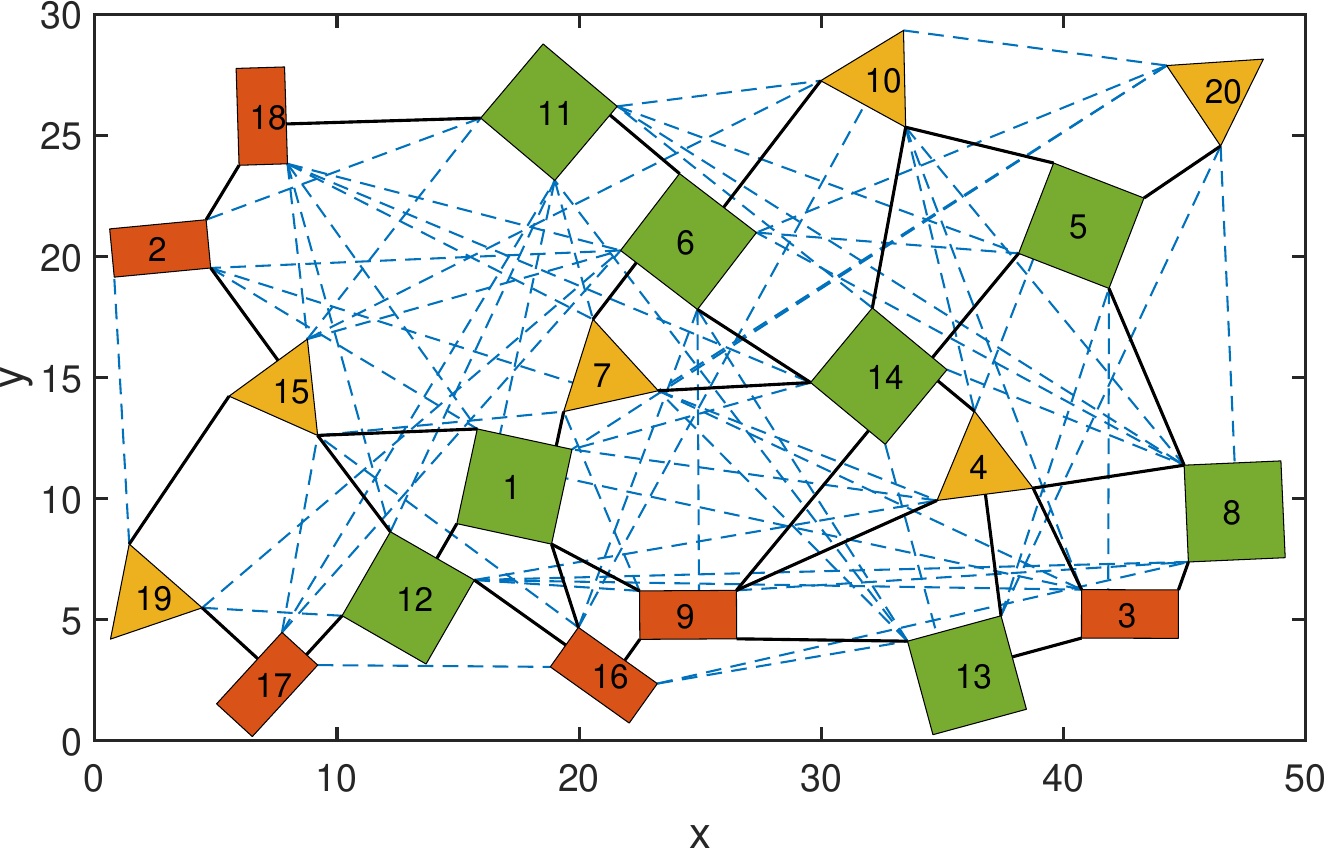}}
    \endminipage  \hfill 
    \caption{Examples of passage detection results with different obstacle distributions. Dashed blue segments are passages after the visibility check. Solid black segments are passages after the extended visibility check. (a) and (b) have the same obstacle side length of one with 20 and 100 obstacles respectively. Dense dashed segments are not plotted in (b) for clarity. (c) 20 obstacles with an obstacle side length of four.}
    \label{Passage Check Different Sizes}
\end{figure*}

\subsection{Coordinated Deformable Path Transfer}
After attaining $\sigma_p^*$, the next step generates rest agent paths. Although $\sigma_p^*$ gets further optimized based on $\sigma_p$, directly transferred paths from $\sigma_p^*$ via (\ref{Combined Path Trasfer}) may still be infeasible in constrained environments. To effectively convert a locally infeasible path into a feasible one, a path-deforming scheme similar to the path-guided optimization avenue in \cite{B. Zhou 2020}, \cite{B. Zhou 2021} is leveraged. The core idea is to use proper reference points to tailor infeasible path parts to feasible paths through path deformation. Firstly, a new path set $\Sigma^*_t$ is regenerated by transferring $\sigma_p^*$ to rest agents as (\ref{Combined Path Trasfer}). Passage intersection check is conducted as before. If adjustment is needed, the reference point for each transferred path is given as
\begin{equation}
\label{Reposition Intersection Points}
   \sigma_{t, j}^*(\eta_{j, i}) = \sigma_p^*(\eta_{p, i}) + r_i (\sigma_{t, j}(\eta_{j, i}) - \sigma_p^*(\eta_{p, i}))
\end{equation}
where $\sigma_{t, j}^*(\eta_{j, i})$ is the fixed reference point. The intersection point of the repositioned $\sigma_p^*$ is fixed as the reference point. This follows the proportional distribution in (\ref{Reference Point Position 2}) to compress the chord. With all these reference points, each transferred path is deformed in the same manner as (\ref{Reposition Pivot Path}).

The deformed transferred path $\sigma_{t,j}^*$ need not be collision-free because obstacle sizes may not be negligible in practice. The key to addressing this is providing more reference points near narrow passages than those on the passage segments. This can be achieved by introducing translated passage segments on obstacle vertices as shown in Fig. \ref{Generalized Passage Segments} and more details are provided in Appendix \ref{App B}. The overall path set generation pipeline is outlined in Algorithm \ref{General Path Set Generation}. The completeness of the entire scheme is ensured by the completeness of the optimal planner backbone for pivot path planning. Only linear time complexity w.r.t. the discrete path node number exists to obtain a transferred path and the resulting paths are strong path homotopic by construction.

\section{Experimental Results}
The proposed path set planning pipeline together with core modules is implemented and tested in various conditions. This section presents comprehensive evaluation results. Core code is updated at \href{https://github.com/HuangJingGitHub/HPSP}{\textit{https://github.com/HuangJingGitHub/HPSP}}.

\subsection{Passage Detection by Extended Visibility Check}
As a key upstream module, passage detection determines passages for the following path set planning. The experiment aims to investigate how the visibility condition and the extended variant affect the resulting passages. Different setups of obstacle shapes, sizes, and densities are tested to enrich passage variations. For each obstacle number equally spaced from $10$ to $100$, the passage number is averaged over $10$ random obstacle distributions (map size: $50 \times 30$) of random shapes (squares, regular triangles, and rectangles with an aspect ratio of $2 {:} 1$) and poses. The obstacle size is controlled by the side length which is set as one here to accommodate more obstacles (see Fig. \ref{Passage Check Different Sizes}). The statistical results in Fig. \ref{Passage Number Statistics Different Obstacle Numbers} show that the combinatorial quadratic increase of the passage number w.r.t. the obstacle number is reduced to significant linear relations by two visibility conditions, which dramatically brings down the valid passage number. Both conditions have coefficients of determination larger than $0.99$ after linear regression. The extended visibility condition, however, has a much smaller passage number increase rate ($\sim 2.1$ vs. $15.0$). The ratios of the passage number using the visibility condition to that using the extended version has a mean of $0.158$, suggesting that only a small fraction of passages remain after further checking the extended visibility.

\begin{figure}[t]
    \minipage{1 \columnwidth}
    \centering
    \subfigure[]{
    \includegraphics[width= 0.8\columnwidth]{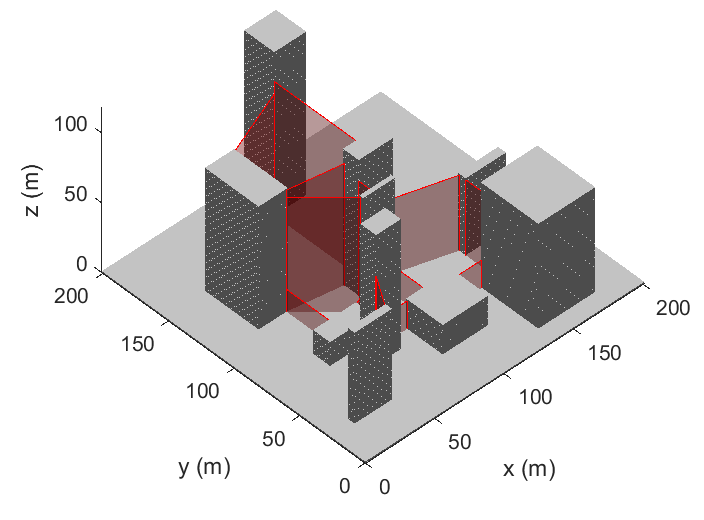}
    \label{3D Map}
    }
    \endminipage \hfill
    \iftrue
    \vspace{0.78 \baselineskip}
    \minipage{1 \columnwidth}
    \centering
    \subfigure[]{
    \includegraphics[width= 0.88 \columnwidth]{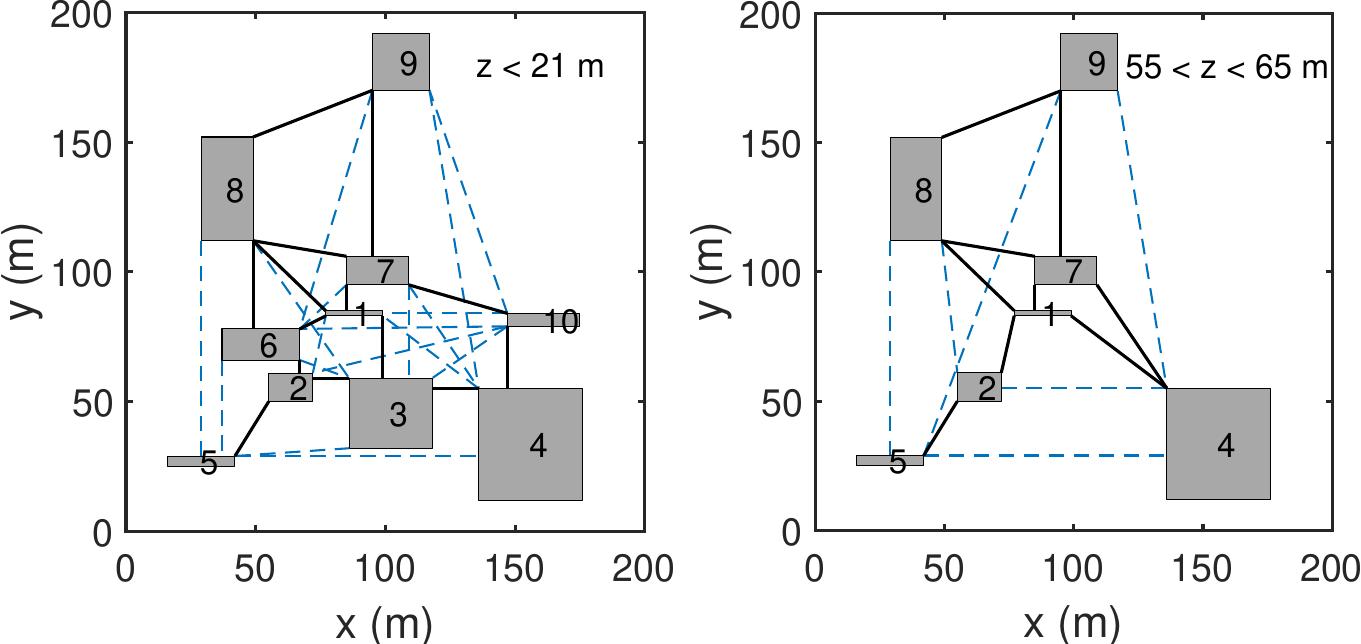}    
    \label{Height Varying Passages for 3D}
    }
    \endminipage \hfill
    \fi
    \caption{3D passage detection example. (a) Valid passages are transparent red planes. 
    (b) Height-varying passage distributions in two height intervals.
    }
    \label{Passage Check Result 3D}
\end{figure}

Next, the obstacle number is fixed as 20, and side lengths are equally spaced from $0.5$ to $5$ to change obstacle sizes. Obstacle distributions are still randomly generated in $10$ tests for each side length. Unlike the pure visibility check, Fig. \ref{Passage Number Statistics Different Obstacle Sizes} indicates that the extended visibility check is not sensitive to obstacle size changes. The passage number via the visibility check decreases significantly and nearly linearly as obstacles expand, but passage numbers after the extended visibility check present small variations. The passage number increases slightly rather than decreases as obstacles get larger. This counterintuitive phenomenon can be attributed to the fact that when obstacle sizes grow, the passage segment length $l(\mathbf{p}_i^*, \mathbf{p}_j^*)$ shrinks twice as fast as the third obstacle $\mathcal{E}_k$'s distance to the passage center $\mathbf{o}_{i,j}$, making $ \mathcal{E}_k \cap \mathcal{R}_{i,j} \neq \emptyset $ easier to meet in (\ref{Extended Visibility Check}), leading to insignificant passage number rises in Fig. \ref{Passage Number Statistics Different Obstacle Sizes}. Predictably, the two conditions' differences will be unnoticeable if obstacle sizes are sufficiently large. Finally, Fig. \ref{Passage Check Result 3D} shows an example of passage detection in a 3D map. The passage distribution varies in height intervals divided by obstacle heights to get a sparse result. The passage distribution can be retrieved efficiently by indexing height as the key.

\subsection{Passage-Aware Optimal Path Planning Results}
This part showcases passage-aware optimal path planning (PAOPP) results. We aim to investigate two major aspects: the influence of cost formulations on planned paths and computational performance differences brought by two visibility checks in path planning. Despite many available efficient planner implementations such as the open motion planning library \cite{I. A. 2012} and the navigation toolbox in MATLAB, it is not straightforward to incorporate passage-related functions and customized costs into existing frameworks due to the lack of related interfaces. RRT$^*$ planner is thus implemented separately with all subroutines in C$++$. Obstacle-related functionalities, including two types of visibility check for passage detection, passage segment positioning, and passage passing check for path segments, are packaged to be invoked readily. These make cost forms and parameters easily configurable when instantiating a planner. 
\begin{figure}[t]
    \minipage{1 \columnwidth}
    \centering
    \subfigure[]{
    \includegraphics[width= 0.91 \columnwidth]{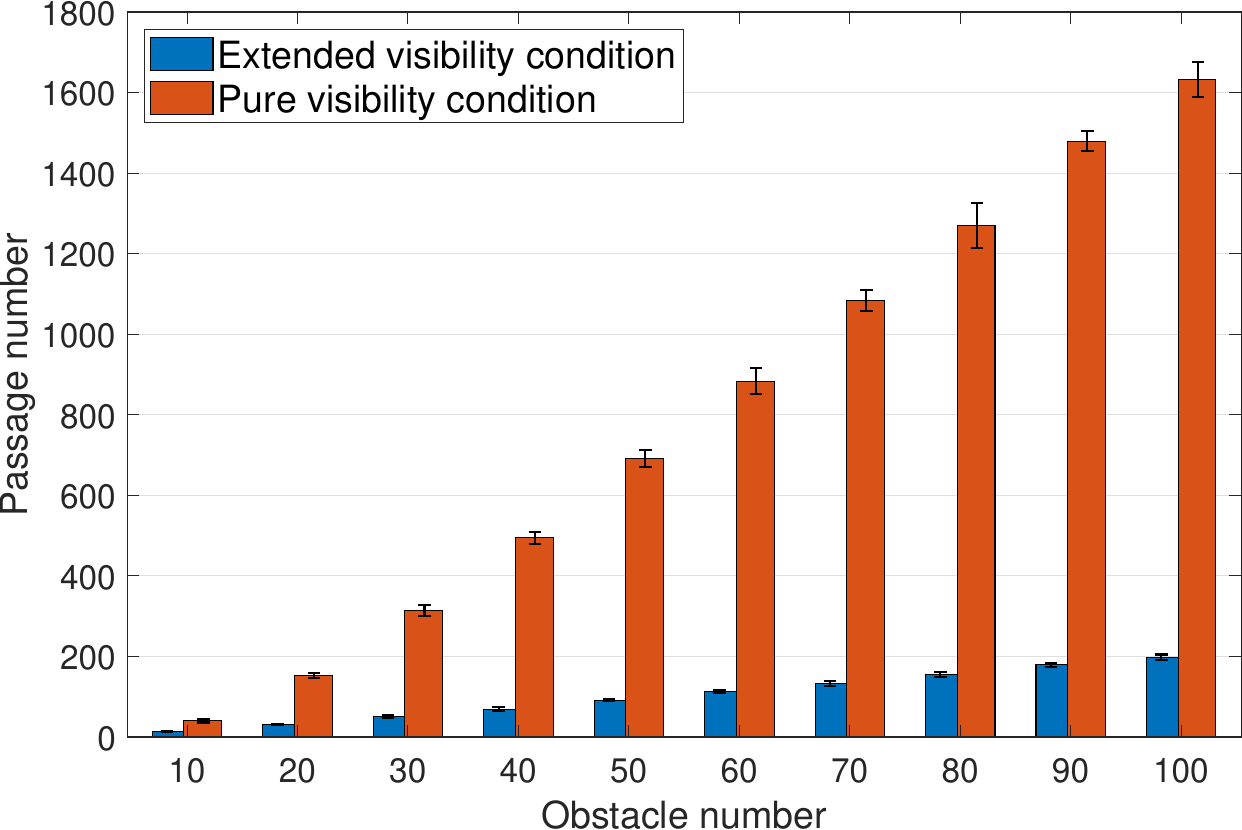}
    \label{Passage Number Statistics Different Obstacle Numbers}
    }
    \endminipage \hfill
    \minipage{1 \columnwidth}
    \centering
    \subfigure[]{
    \includegraphics[width= 0.91 \columnwidth]{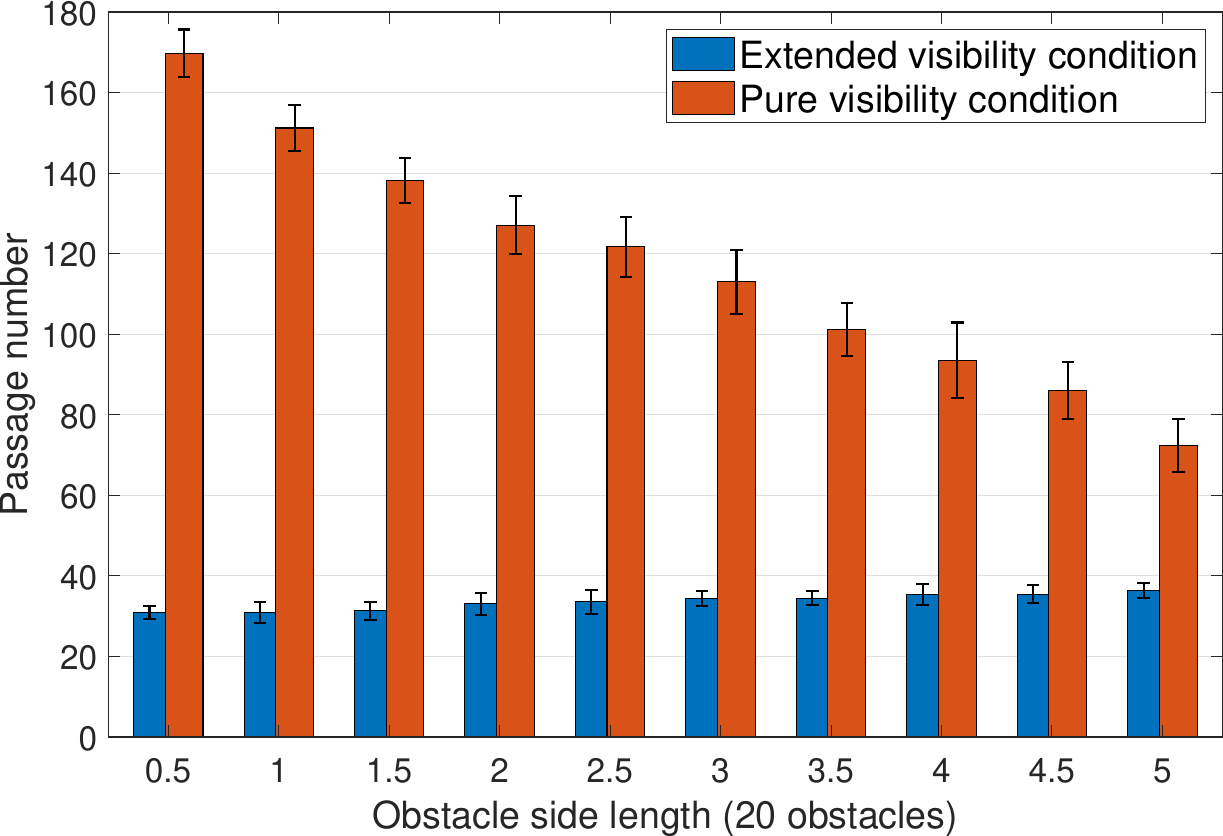}    
    \label{Passage Number Statistics Different Obstacle Sizes}
    }
    \endminipage \hfill
    \caption{Statistical results of passage numbers detected by two check conditions in different setups. (a) Passage numbers with different numbers of obstacles (obstacle side length is one). (b) Passage numbers with the same number of obstacles but different obstacle sizes.}
    \label{Passage Number Statistics}
\end{figure}

\begin{figure*}[t]
    \minipage{2 \columnwidth}
    \centering
    \subfigure[$M = 10$] {
    \includegraphics[width= 0.31 \columnwidth]{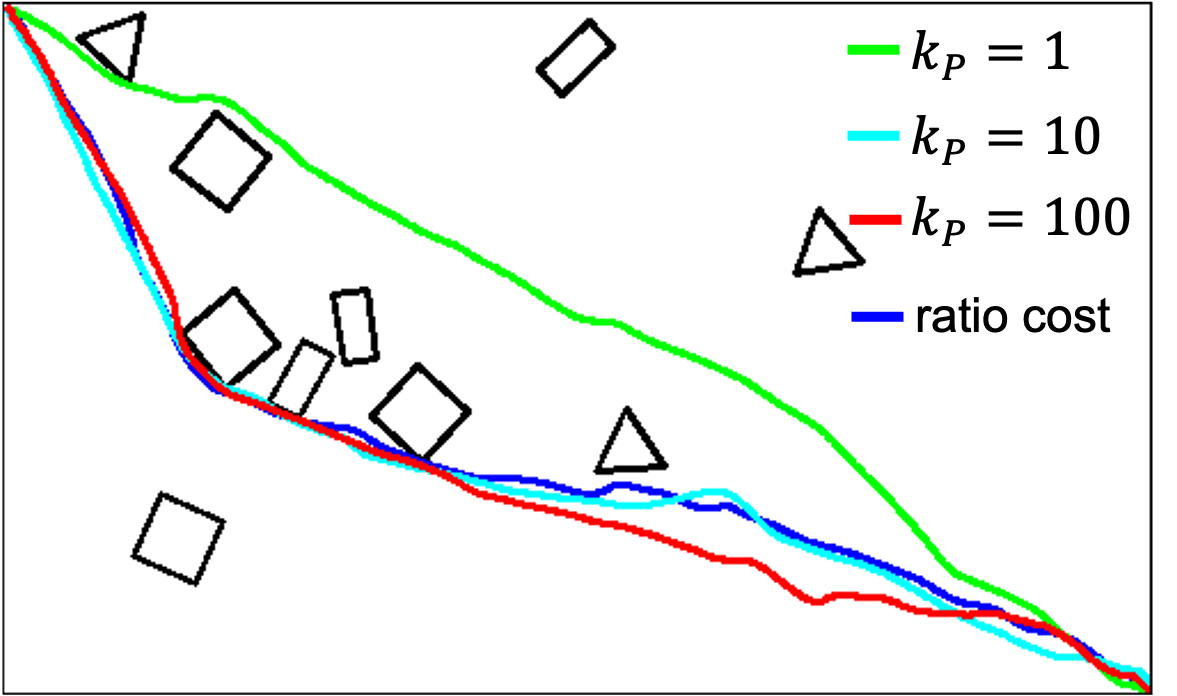}}
    \subfigure[$M = 20$] {
    \includegraphics[width= 0.31 \columnwidth]{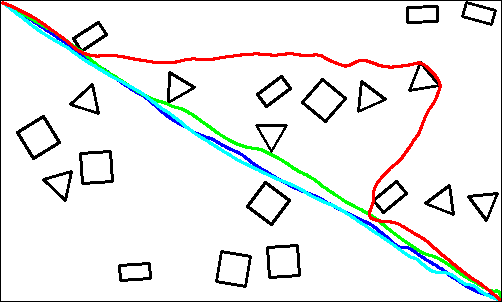}}
    \centering
    \subfigure[$M = 30$] {
    \includegraphics[width= 0.31 \columnwidth]{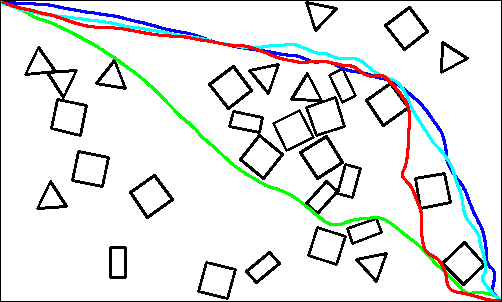}}
    \endminipage  \hfill 
    \minipage{2 \columnwidth}
    \centering
    \subfigure[$M = 40$] {
    \includegraphics[width= 0.31 \columnwidth]{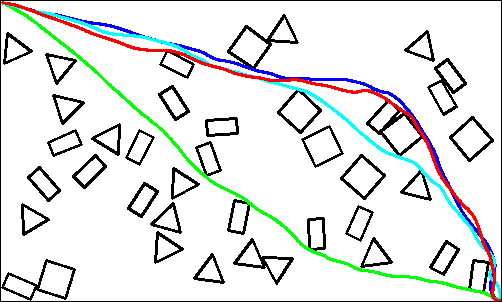}}
    \subfigure[$M = 50$] {
    \includegraphics[width= 0.31 \columnwidth]{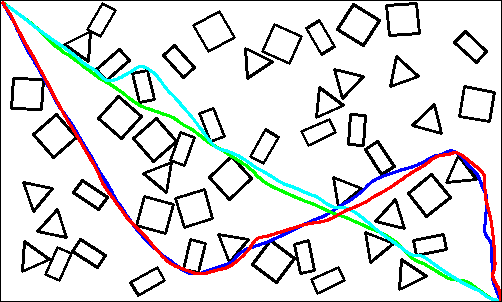}}
    \centering
    \subfigure[$M = 60$] {
    \includegraphics[width= 0.31 \columnwidth]{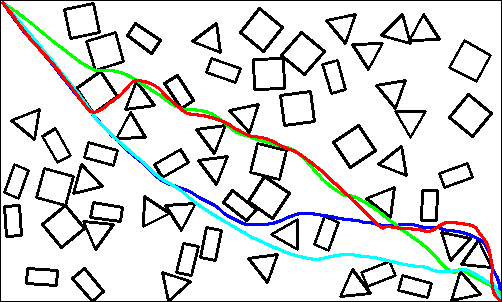}}
    \endminipage  \hfill     
    \caption{Path planning results using different cost formulations. The obstacle number in the environment ranges from $10$ to $60$. Blue paths correspond to the cost in (\ref{Cost Function for Passage Passing}). Green, cyan, and red paths correspond to the cost in (\ref{Weighted Cost Function for Passage Passing}) with $k_P = 1, 10$, and $100$, respectively.} 
    \label{Path Planning With Different Costs}
\end{figure*}

As depicted in Fig. \ref{Path Planning With Different Costs}, different cost formulations are tested in path planning: the ratio cost in (\ref{Cost Function for Passage Passing}) and the weighted cost in (\ref{Weighted Cost Function for Passage Passing}) with different weights ($k_P = 1, 10$, and $100$ respectively to change the preference). Various numbers of obstacles are randomly distributed. Passage detection is conducted using the extended visibility check and environment boundaries are treated as extra obstacles. The planning problem is constant across tests to find an optimal path from the top left to the bottom right corner as in Fig. (\ref{Path Planning With Different Costs}). It is observed that paths can vary with cost choices. For the weighted cost, the path length Len$(\sigma)$ dominates the cost when $k_P = 1$. Thus, the resulting paths (green paths) prioritize minimizing Len$(\sigma)$, similar to the most typical setup in optimal path planning. In comparison, when $k_P = 100$, the minimum traversed passage width $f_P(\sigma)$ largely determines the cost. Planned paths (red paths) try to avoid going through narrow passages. In between, $k_P = 10$ balances the two items shown by cyan paths. The value range of $k_P$ to effectively alter two factors' precedence is problem-related as $k_P$ varies to make Len$(\sigma)$ and $k_P f_P(\sigma)$ comparable. Overall, the ratio cost (paths are in blue) behaves similarly to adopting a moderate $k_P$ in the weighted cost, making it a balanced cost in most cases.

To quantitatively measure how the extended visibility check improves planning efficiency, passages are identified by two check conditions respectively for PAOPP, i.e., PAOPP-pure and PAOPP-ext. Planning efficiency is gauged by the planning time with a maximum valid sample number $N = 10 $k. The path cost is weighted ($k_P = 10$) and the start and goal remain unchanged. All computations are run on a PC with Ubuntu 18.04, Intel Core i9-7980XE CPU$@\SI{2.60}{\GHz} \times 36$, and 64 GB of RAM, with no specifications for acceleration. For the same problem, paths found with two passage check results have the same passage traversal list and similar path costs. Thus, they can be regarded as equivalently optimal paths. See Appendix \ref{App D} for theoretical analyses of two checks' equivalence in PAOPP. The planning time differs much in Table \ref{Tab: Planning Time}. For each obstacle number (the side length is three), 10 different planning tests are run. The planning time is shown to reduce $\SI{58.7}{\percent}$ on average. Roughly, the extended visibility check helps save more than half the planning time. Note that despite the significant time saving, its reduction is not as dramatic as the passage number reduction in Fig. \ref{Passage Number Statistics Different Obstacle Numbers} because passage-related operations only take a fraction of planning computations. 

\begin{table}[t]
\centering
    \caption{Average Time of Passage-Aware Optimal Path Planning and Max-Clearance Path Planning Tests \textnormal{(ms)}}
    \label{Tab: Planning Time}
    \centering
    \begin{tabularx}{1 \columnwidth}{ l Y Y Y Y Y Y}
    \toprule
        Obstacle Num. &$10$  &$20$  &$30$  &$40$  &$50$  &$60$ \\
        \hline         
        MC-time &$327 \times 9.0$  &$619 \times 9.0$  &$740 \times 9.0$  &$916 \times 9.0$  &$1361 \times 8.9$  &$1345 \times 8.7$  \\   
        MC-sample &868  &1440 &2087 &2594 &3627 &4223 \\ 
        PAOPP-pure &1995  &4172  &6342  &8312  &10161  &11405  \\ 
        PAOPP-ext &\textbf{882}  &\textbf{1640} &\textbf{2401} &\textbf{3331} &\textbf{4212} &\textbf{5119} \\ 
    \bottomrule
    \end{tabularx}
\end{table}

Next, PAOPP is further compared to the conventional max-clearance path planning (MCPP) method. MCPP finds the shortest path with the maximum clearance (MC) to all obstacles and represents a classical avenue to balance path's accessible free space and length. To get the MC, a binary search for MC in planning is performed as in Algorithm \ref{Binary-Search MCPP}. The lower and upper bounds are half the minimum and maximum passage width respectively. For completeness, passage widths are detected by the pure visibility check. Two failure criteria for an MC value exist: return false if the planner fails to find a path after a given time (MCPP-time) or a given sample number (MCPP-sample). In Table \ref{Tab: Planning Time}, MCPP-time reports the one-round planning time when MC is found within an error of $0.5$ $\times$ the total planning trail number. MCPP-sample reports the total planning time and the max sample number is $N_{MC} = 20$k. MCPP-time turns out to be the slowest. Due to the fast planning failure check, MCPP-sample is faster than PAOPP with sparse passages by $\SI{13.4}{\percent}$ on average. However, its inherent drawbacks are that the planned result is not adjustable and its performance deteriorates if $N_{MC}$ is large. See Appendix \ref{App C} for more implementation details and results.

\subsection{Path Set Generation Results}
\begin{figure*}[t]
    \minipage{2 \columnwidth}
    \centering
    \subfigure[] {
    \includegraphics[width= 0.31 \columnwidth]{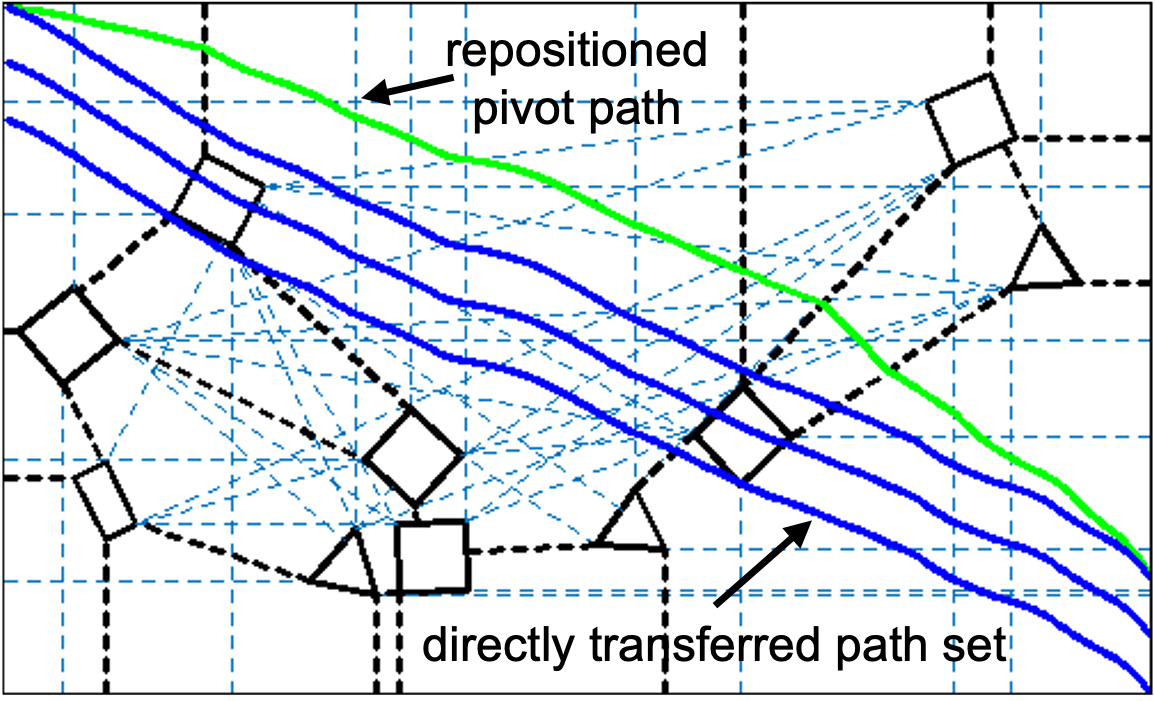}
    \label{Path Set Generation Reposition a}}
    \subfigure[] {
    \includegraphics[width= 0.31 \columnwidth]{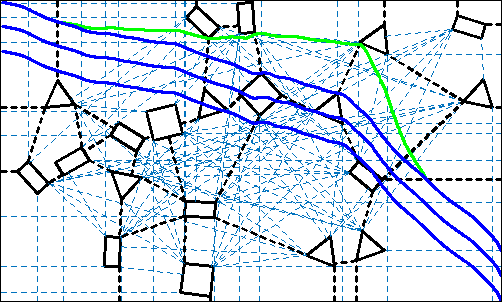}
    \label{Path Set Generation Reposition b}}
    \subfigure[] {
    \includegraphics[width= 0.31 \columnwidth]{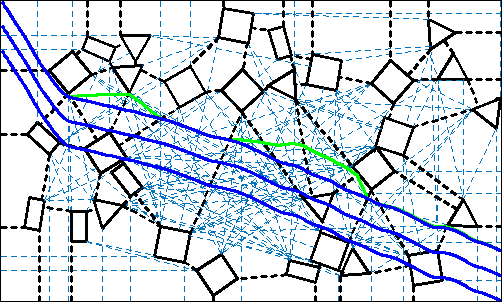}
    \label{Path Set Generation Reposition c}}
    \endminipage  \hfill 
    \minipage{2 \columnwidth}
    \centering
    \subfigure[] {
    \includegraphics[width= 0.31 \columnwidth]{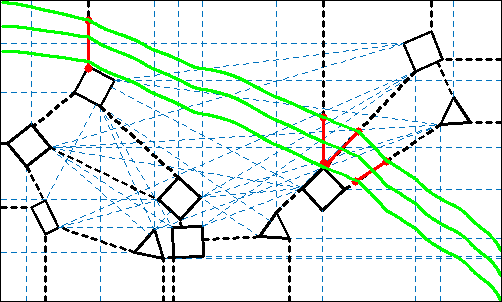}
    \label{Path Set Generation Reposition d}}
    \subfigure[] {
    \includegraphics[width= 0.31 \columnwidth]{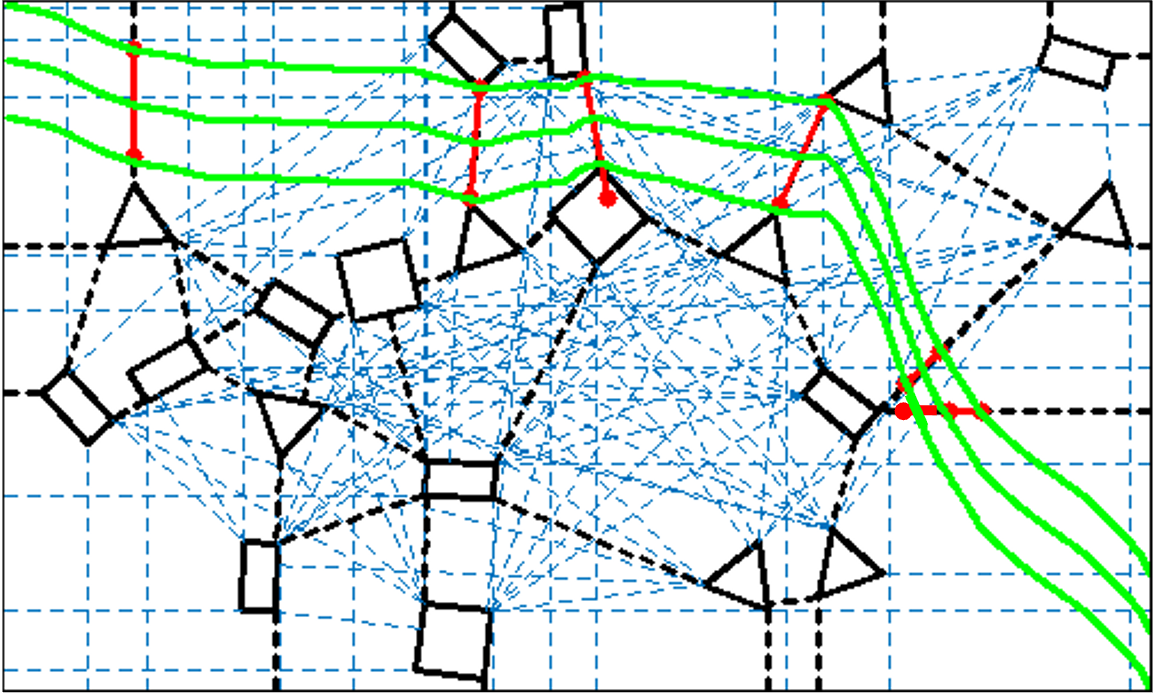}
    \label{Path Set Generation Reposition e}}
    \subfigure[] {
    \includegraphics[width= 0.31 \columnwidth]{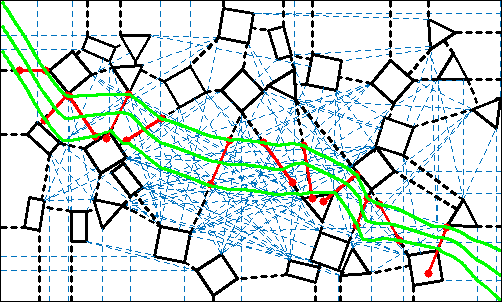}
    \label{Path Set Generation Reposition f}}
    \endminipage  \hfill     
    \caption{Path set generation results. Black dashed lines are passages after the extended visibility check. Light blue dashed lines are passages after the pure visibility check. (a)-(c) show the directly transferred path sets in blue and the repositioned pivot paths in green. (d)-(f) show corresponding path sets obtained via deformable path transfer from repositioned pivot paths. Red segments depict the chords of $\Sigma^*_t$ transferred from the repositioned $\sigma_p^*$.} 
    \label{Path Set Generation}
\end{figure*}

Built on the modules above, the set generation scheme is implemented and tested. Given the initial $S_0$ and final $S_d$, a pivot path is first planned via PAOPP. $k_P = 10$ is utilized in pivot path planning for a balanced cost. The top row in Fig. \ref{Path Set Generation} illustrates the directly transferred path set $\Sigma_t$ (in blue) and the repositioned pivot path $\sigma_p^*$ (in green). Though the planned pivot path $\sigma_p$ goes through passages associated with large free space, directly transferred paths collide with obstacles easily. This is because the passage passing positions of $\sigma_p$, namely intersections with passages, are obtained to minimize the path length, which commonly makes $\sigma_p$ located in obstacles' vicinity. Thereby, repositioning $\sigma_p$ is required while preserving its passed passages to better accommodate transferred paths. In experiments, the geometrical approach for chord determination is utilized. Reference points on passage segments are given from chords' locations and $\sigma_p$ is deformed piecewise to $\sigma_p^*$ as (\ref{Reposition Pivot Path}).  

It is reinforced that filtering out invalid passages is not only for computational efficiency but also necessary. Dense passages after pure visibility check make path set intersections overly redundant. The extended variant attains a sparse distribution of passages more suitable for repositioning $\sigma_p$ and deformable path transfer. The repositioned $\sigma_p^*$ is in a configuration more likely to lead to feasible transferred paths. In less restrictive cases with fewer obstacles like Fig. \ref{Path Set Generation Reposition a} and Fig. \ref{Path Set Generation Reposition d}, paths directly transferred from $\sigma_p^*$ are feasible. In general, transferred paths from $\sigma_p^*$ may collide with obstacles, and coordinated deformable path transfer is required. As in Fig. \ref{Path Set Generation Reposition f}, there exist narrow passages that cannot accommodate all directly transferred paths. Coordinated reference points are given for each path to guide path deformation, making the path set go through narrow passages freely. In scenarios where obstacles are dense and of significant sizes, translated passage segments can be introduced to density reference points for feasible deformed paths. See Appendix \ref{App B} for more examples with translated passages.
\begin{table}[t]
\centering
    \caption{Average Path Set Planning Time Using Path Set Transfer and Separately Planning Method \textnormal{(ms)}}
    \label{Tab: Path Set Planning Time}
    \centering
    \begin{tabularx}{1 \columnwidth}{ l Y Y Y Y Y Y}
    \toprule
        Path Number  &$3$  &$6$  &$9$  &$12$  &$15$  &$18$ \\
        \hline
        SP ($M = 10$) &4442  &9501  &14653  &20083  &25393  &31105  \\          
        PT ($M = 10$) &\textbf{840}  &\textbf{882}  &\textbf{845}  &\textbf{879}  &\textbf{863}  &\textbf{867}  \\   
        \hline
        SP ($M = 20$) &5809  &11972  &17743  &24630  &28956  &33411  \\          
        PT ($M = 20$) &\textbf{1660}  &\textbf{1681}  &\textbf{1653}  &\textbf{1665}  &\textbf{1661}  &\textbf{1658}  \\   
        \hline
        SP ($M = 30$) &7242  &13259  &21332  &28087  &34020  &40727  \\          
        PT ($M = 30$) &\textbf{2480}  &\textbf{2433}  &\textbf{2498}  &\textbf{2450}  &\textbf{2452}  &\textbf{2272}  \\           
    \bottomrule
    \end{tabularx}
\end{table}

For efficiency evaluation, the benchmark method of separately planning (SP) each path is compared with the proposed scheme based on path transfer (PT). In the SP method, $\sigma_p$ is also first planned using PAOPP. Then the rest paths are planned separately with the homotopy constraint to $\sigma_p$. Specifically, samples are restricted to be close to $\sigma_p$ and are obstacle-free to $\sigma_p$ in the neighborhood. Table \ref{Tab: Path Set Planning Time} outlines the average path set planning time when obstacle number and path number vary in 10 random setups. As is observed, the SP method has a significantly larger time cost that increases nearly linearly with the path number. Conversely, the PT method time is not sensitive to the path number at all. Its dominant cost solely comes from $\sigma_p$ planning in PAOPP. Other operations in PT (e.g., pivot path reposition, path transfer) are linear to the path resolution, i.e., path node number, which is almost negligible compared to the cost of invoking the planner once. Moreover, coordination constraints are hard to impose in SP. Thus, the PT method is much more advantageous in homotopic path set planning. See Appendix \ref{App C} for more results.

\subsection{Application Case Study}
\begin{figure*}[t]
    \minipage{2 \columnwidth}
    \centering
    \subfigure[] {
    \label{Case Study Robot View a}
    \includegraphics[width= 0.235 \columnwidth]{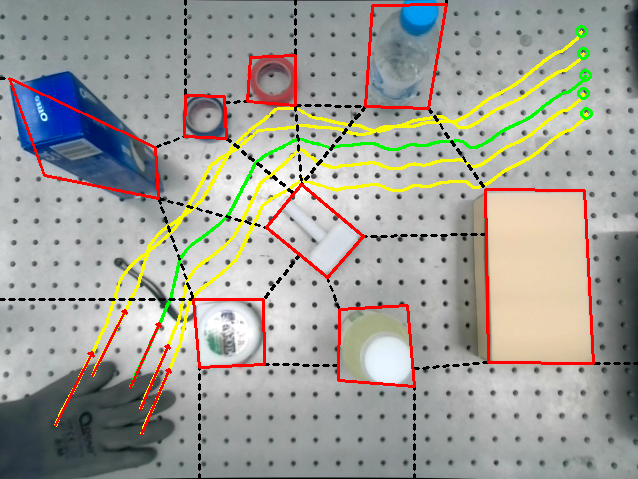}}
    \subfigure[] {
    \label{Case Study Robot View b}
    \includegraphics[width= 0.235 \columnwidth]{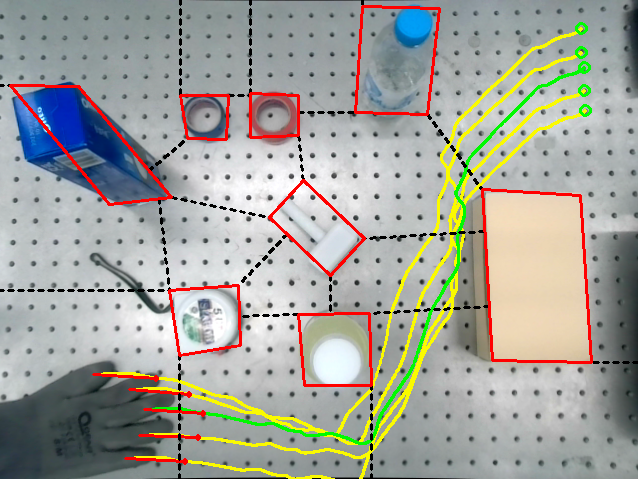}}
    \subfigure[] {
    \label{Case Study Robot View c}
    \includegraphics[width= 0.235 \columnwidth]{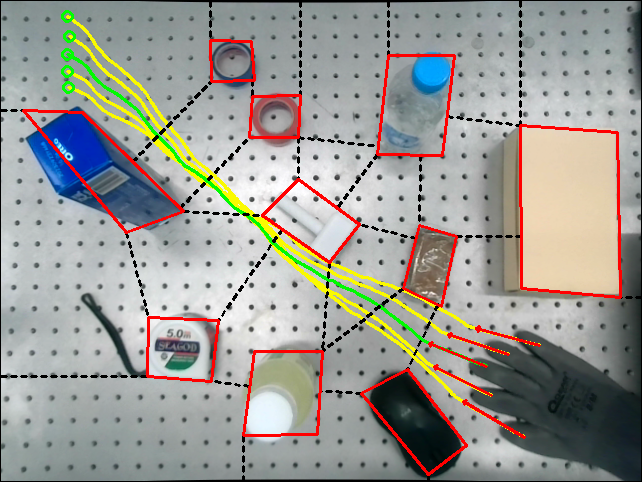}}
    \subfigure[] {
    \label{Case Study Robot View d}
    \includegraphics[width= 0.235 \columnwidth]{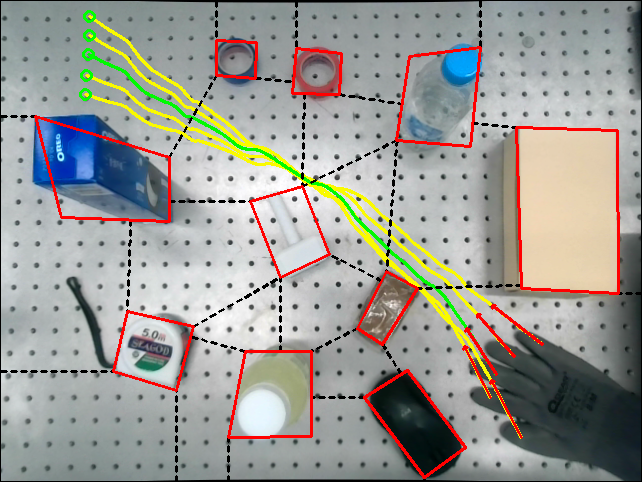}}
    \endminipage  \hfill    
    \caption{Path set planning in DO manipulation site with random daily objects (bottles, boxes, tapes, etc.). There are eight objects in (a) and (b), ten in (c) and (d). Black dashed lines are detected passages. Planned path sets for the glove comprise yellow and green paths (pivot paths).}
    \label{Case Study Robot View}
\end{figure*}

The last part showcases some common applications of path set planning in robot manipulation and navigation. The first kind of task is DO path planning in complex environments, similar to \cite{J. Huang 2023 TRO}, but in much more cluttered setups. As shown in Fig. \ref{Case Study Robot View}, a manipulation site is emulated with various objects randomly placed on the table. The robot needs to move a deformable grip glove on this table while avoiding collisions. Due to the restriction of tall objects like vertically placed boxes and bottles, manipulation is confined near the table surface, for which a path reference of the glove is required. To do this, several keypoints are picked on the glove to depict the glove state. In Fig. \ref{Case Study Robot View}, these points are mainly distributed on fingers and the middle one acts as the pivot by (\ref{Pivot Selection}). The target $S_d$ is vertically aligned. Obstacles are segmented as bounding quadrangles and path set planning is conducted in the image space. Fig. \ref{Case Study Robot View} demonstrates different test results. As can be seen, the planned path sets have wide free space along them with sufficiently short lengths, making them good path references for glove movement. With planned path sets, subsequent manipulation can be conducted as a constrained path set tracking problem as in \cite{J. Huang 2023 TRO}.
\begin{figure*}[t]
    \minipage{2 \columnwidth}
    \centering
    \subfigure[] {
    \label{Case Study Swarm a}
    \includegraphics[width= 0.235 \columnwidth]{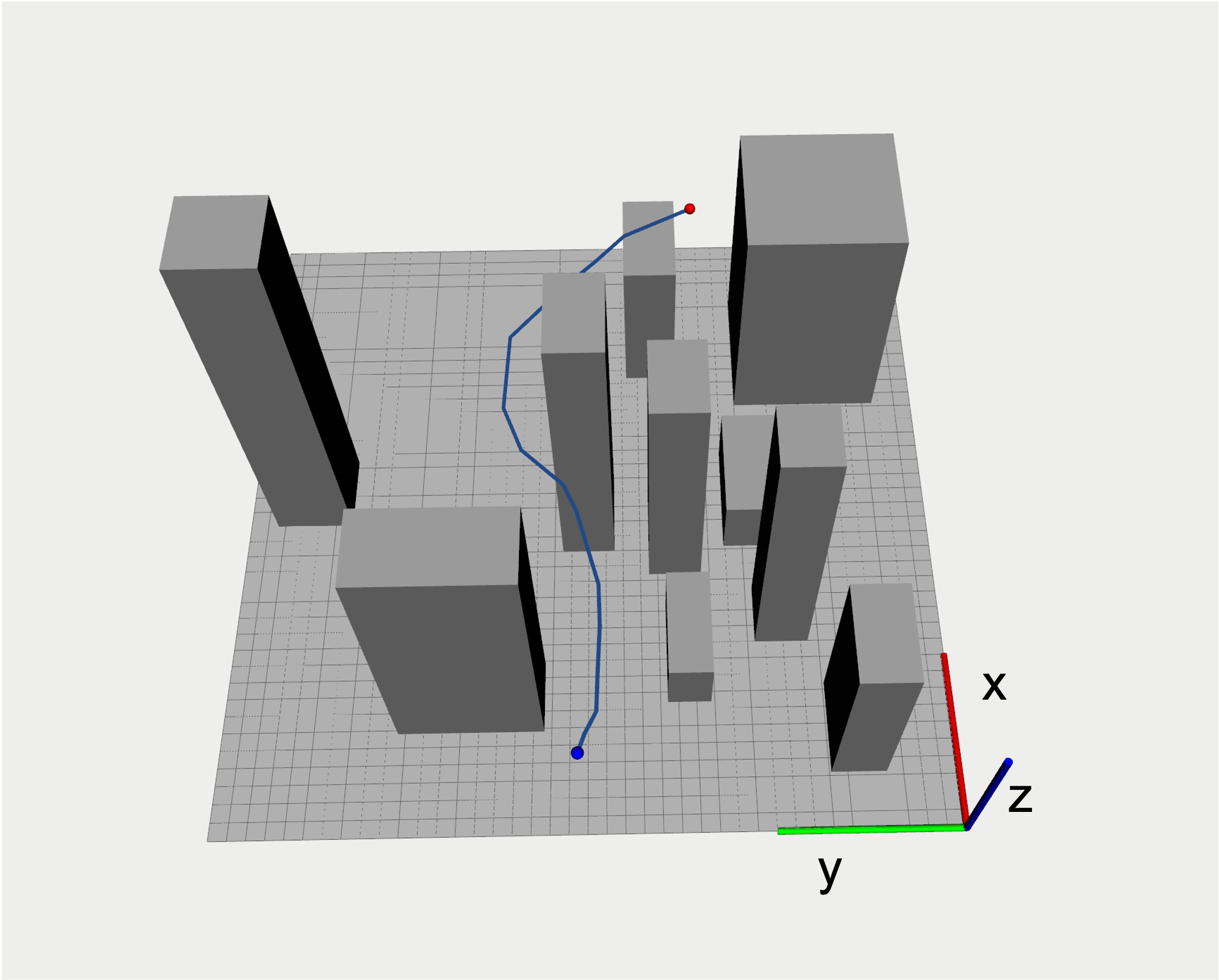}}
    \subfigure[] {
    \label{Case Study Swarm b}
    \includegraphics[width= 0.235 \columnwidth]
    {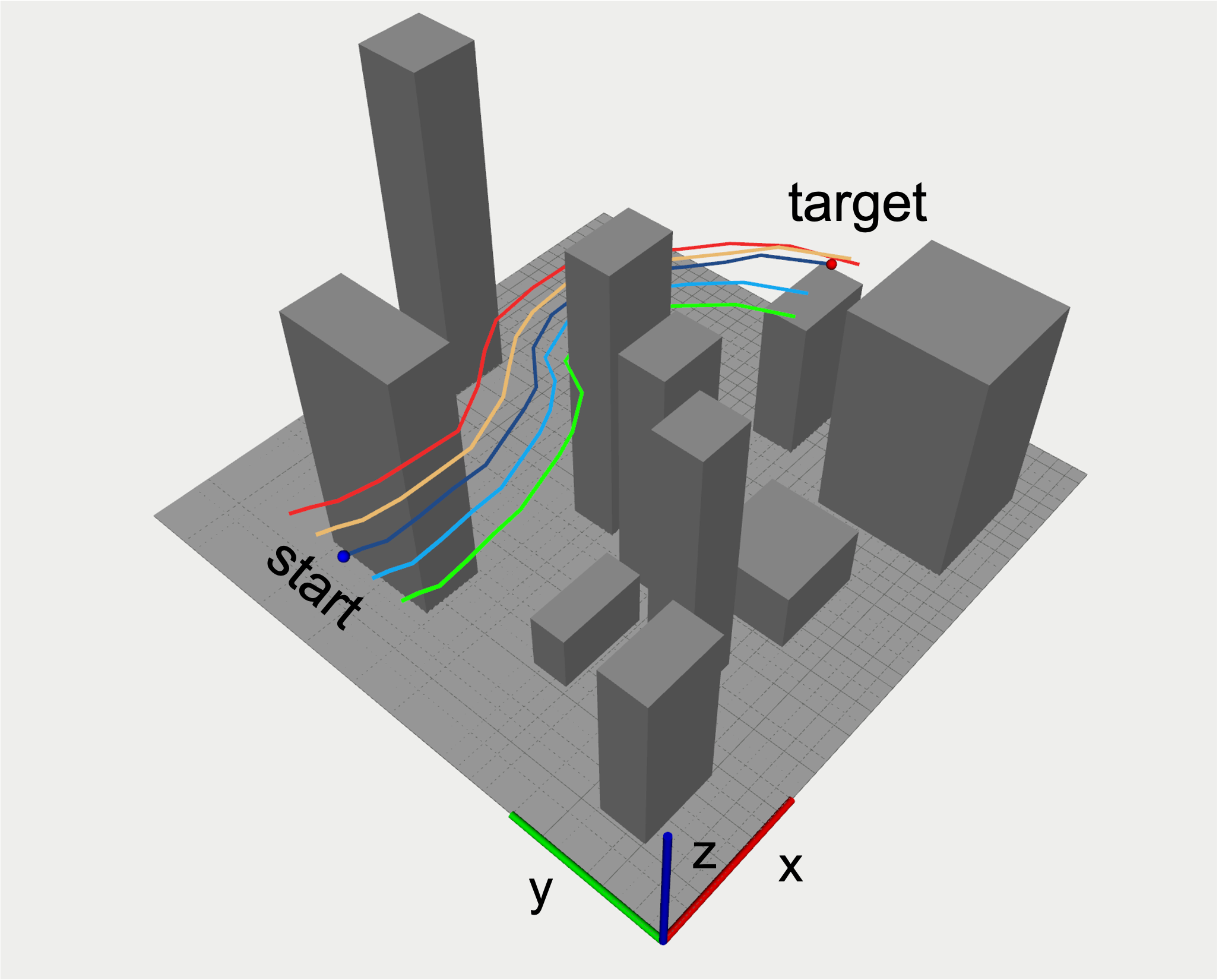}}
    \subfigure[] {
    \label{Case Study Swarm c}
    \includegraphics[width= 0.235 \columnwidth]{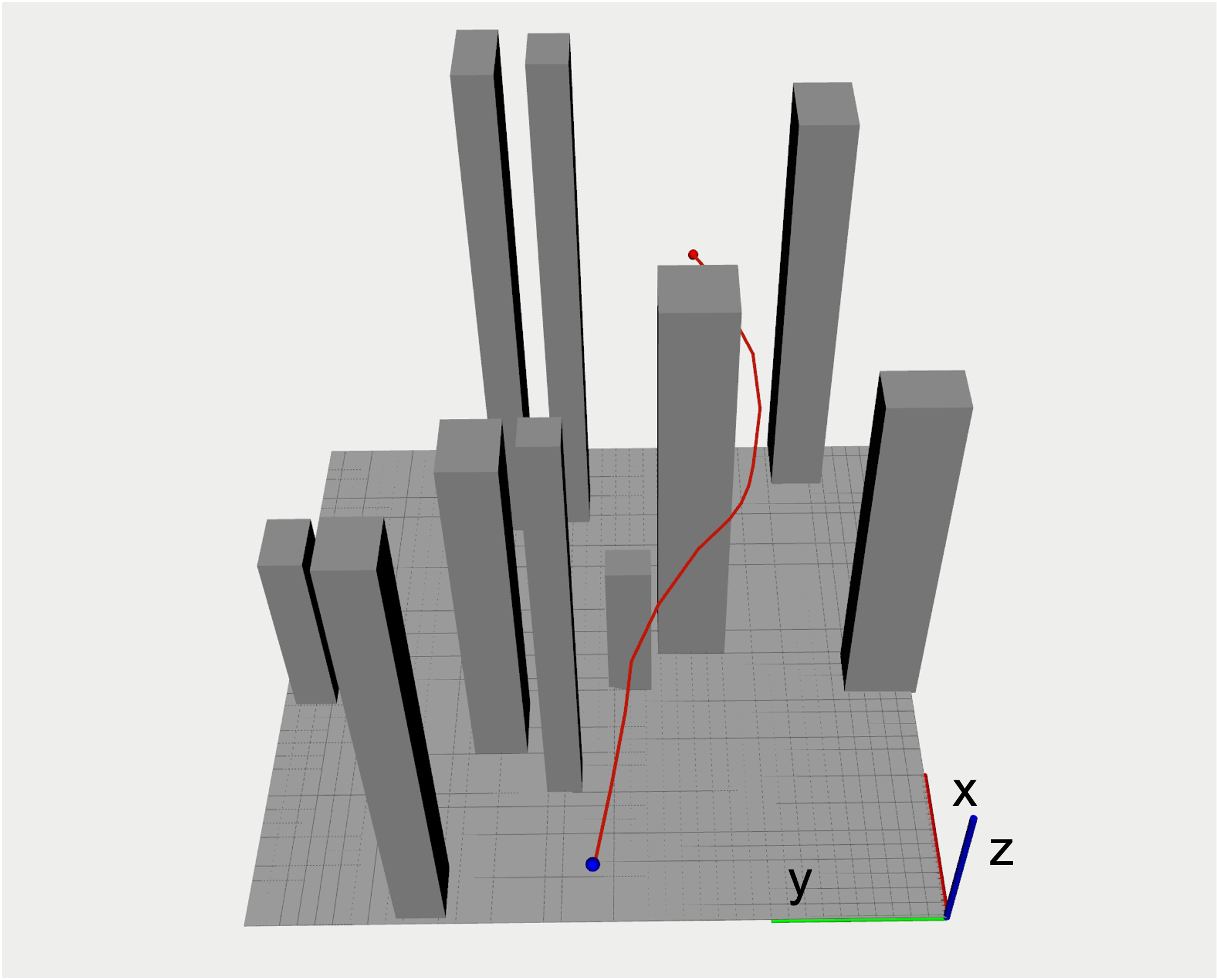}}
    \subfigure[] {
    \label{Case Study Swarm d}
    \includegraphics[width= 0.235 \columnwidth]{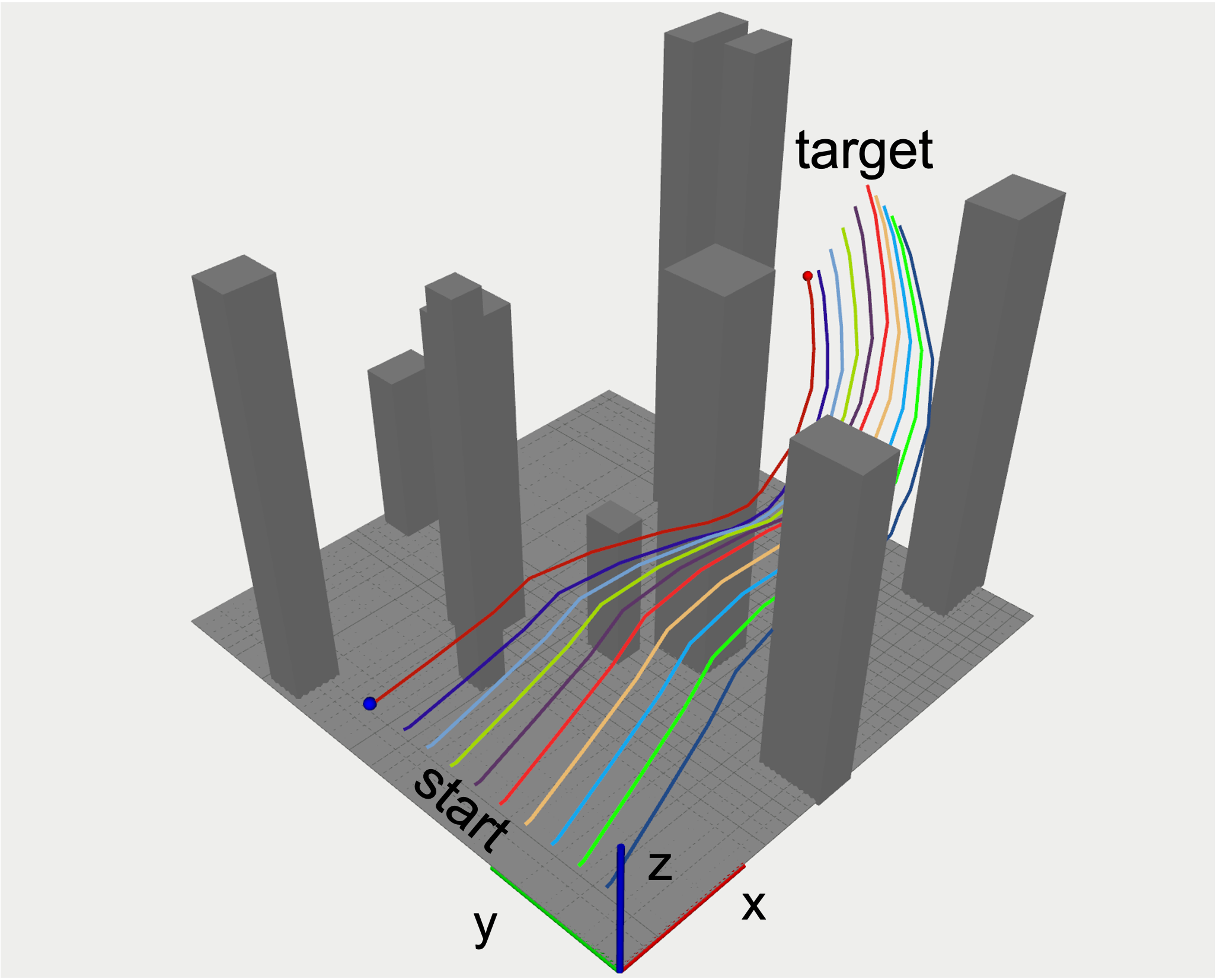}}
    \endminipage  \hfill    
    \caption{Path set planning examples for swarms ($K = 5, 10$) in 3D maps. (a), (c) Passage-aware optimal path planning returns pivot paths with large accessible free space. (b), (d) Corresponding final path sets after pivot path repositioning and coordinated deformable path transfer.}
    \label{Case Study Swarm}
    \end{figure*}

The second task is swarm navigation in 3D maps. In Fig. \ref{Case Study Swarm}, aerial robot swarms ($K = 5, 10$) move as a team in city environments where buildings of various sizes and heights are populated. To reach the target, the swarm needs to fly through narrow spaces among buildings. Robots are not allowed to fly subsequently as a queue or vertically aligned when passing narrow space and they are initially aligned horizontally in a line. The target is a different and distant formation. Robots' positions change in 3D, including the flight height. Passage detection is first performed to establish the passage distribution in height intervals formed by buildings. The pivot can be selected arbitrarily and its planned path selects wide passages in Fig. \ref{Case Study Swarm a} and Fig. \ref{Case Study Swarm c}. Then pivot path repositioning and coordinated deformable path set transfer are performed to generate the final path set in Fig. \ref{Case Study Swarm b} and Fig. \ref{Case Study Swarm d}. Path deformation on passages is restricted in the $x$-$y$ plane. Since robots' initial and target positions are vertices of convex hulls, methods in \cite{P. Mao 2023} will have to plan paths for each robot with no optimization of accessible free space along paths.

\section{Conclusion}
This paper presented a systematic pipeline of homotopic path set planning for substantial robotics applications. The extended visibility check was first proposed to attain sparse passage distributions over pure visibility check. Passage-aware optimal path planning compatible with sampling-based optimal planners was designed with adjustable path costs. It could balance the path length and accessible free space more flexibly over clearance-based planning methods. Path set generation based on path transfer was then proposed. Techniques such as proportional reference point distribution, geometrical chord determination, and translated passage segments provided a better guarantee of the resulting paths' feasibility and coordination. The scheme's effectiveness was validated in different experiments. 
Future work includes more detailed passage descriptions as an area or space. Considering dynamic environments and imposing constraints among agents such as reducing path intersections are also important directions.


\clearpage
\appendices
\renewcommand{\theequation}{A\arabic{equation}}
\setcounter{equation}{0} 
\section{Chord Determination via Path Set's Local Geometry}
\label{App A}
\begin{figure}[t]
    \minipage{1 \columnwidth}
    \centering
    \includegraphics[width= 0.95 \columnwidth]{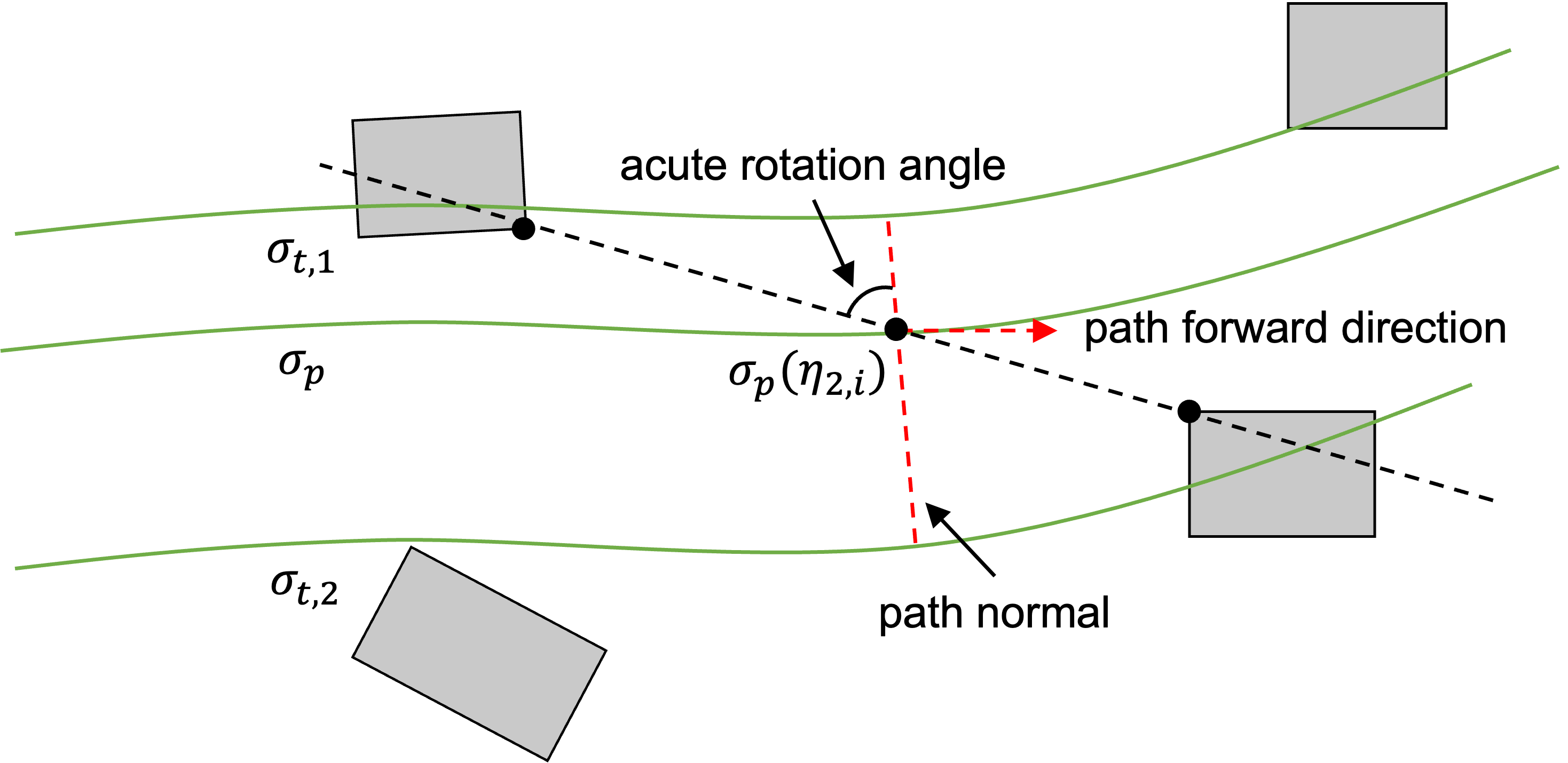}
    \endminipage \hfill
    \caption{In this example, the chord on the passage line poorly characterizes how paths should be moved. Paths' intersections on the normal line of the pivot path better describe the path set's local geometry.}
    \label{Tube Chord Concept}
\end{figure}

The chord obtained via the intersection check of the passage line and the path set is not always informative for repositioning the path set. Regarding the path set as a tube, a chord makes a good reference when it aligns well with the cross-section of the tube as exemplified in Fig. \ref{Tube Chord Concept}. To better describe how the path set passes a passage, instead of intersections with the passage line, the cross-section nearby can be utilized. Specifically, the intersection between the pivot path $\sigma_p$ and passage $P_{\sigma_p}(1, i)$ is $\sigma_p(\eta_{p, i})$. The intersection $\Sigma_t \cap \mathcal{N}(\sigma_p, \eta_{p, i})$ is first found, where $\mathcal{N}(\sigma_p, \eta_{p, i})$ represents the pivot path's normal line at $\sigma_p(\eta_{p,i})$ available by the local path's geometrical properties. Similarly to (\ref{Chord Length}), the chord on  $\mathcal{N}(\sigma_p, \eta_{p, i})$ has the following length form
\begin{equation}
\label{Normal Chord Length}
    \| \Sigma_t \cap \mathcal{N}(\sigma_p, \eta_{p, i}) \|_2 = \max_{1 \leq k, j \leq K} \| \sigma_{t, k}^N(\eta_{k, i}) - \sigma_{t, j}^N(\eta_{j, i}) \|_2
\end{equation} 
where $\sigma_{t, k}^N(\eta_{k, i})$ is the intersection point between $\mathcal{N}(\sigma_p, \eta_{p, i})$ and the transferred path $\sigma_{t,k}$. Since paths' shifts are on the passage. Then $\Sigma_t \cap \mathcal{N}(\sigma_p, \eta_{p, i})$ is reflected on $P_{\sigma_p}(1, i)$, which can be done by rotating $\Sigma_t \cap \mathcal{N}(\sigma_p, \eta_{p, i})$ about the point $\sigma_p(\eta_{p, i})$ to $P_{\sigma_p}(1, i)$. The rotation direction is the acute angle between $P_{\sigma_p}(1, i)$ and $\mathcal{N}(\sigma_p, \eta_{p, i})$ to preserve the relative distribution of intersection points. Roughly speaking, looking in the pivot path's forward direction at $\sigma_p(\eta_{p, i})$, if an intersection point $\sigma_{t, k}^N(\eta_{k, i})$ is on the left (right) side of $\sigma_p(\eta_{p, i})$ on $\mathcal{N}(\sigma_p, \eta_{p, i})$. It should be on the left (right) side of $\sigma_p(\eta_{p, i})$ after rotation on $\mathcal{P}_{\sigma_p}(1, i)$. The chord obtained by rotation now depicts the intersection point distribution on the passage and the following steps keep the same.

\section{Adding Translated Passage Segments on Obstacles}
\label{App B}
\begin{figure}[t]
    \minipage{1 \columnwidth}
    \centering
    \includegraphics[width= 0.66 \columnwidth]{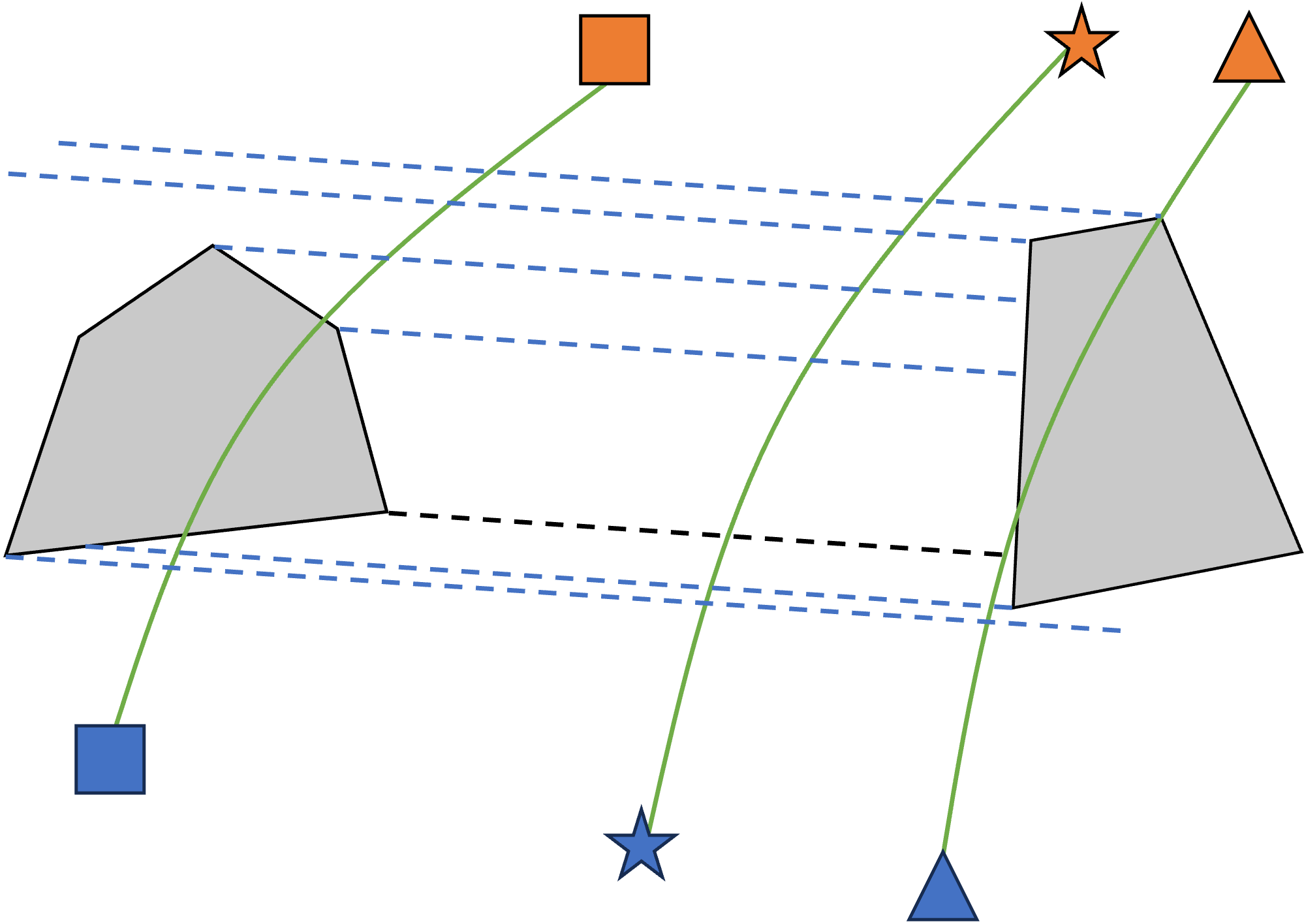}
    \endminipage \hfill
    \caption{Transferred paths from the repositioned $\sigma_p^*$ may be infeasible. The shown paths shifted left from the setup in Fig. \ref{Repositioning Scenario} as $\sigma_p$ is shifted left to $\sigma_p^*$, but infeasible paths exist. Blue dashed lines are translated passage segments starting from obstacle vertices.}
    \label{Generalized Passage Segments}
\end{figure}

Adding translated passage segments aims to provide more references for path collision avoidance in obstacle-dense setups. For example, $\sigma_{t, 3}$ in Fig. \ref{Repositioning Scenario} is deformed to shift left. If the only reference point on $P_{\sigma_2}(1, i)$ is placed too close to the right obstacle, the repositioned path is still in collision and this is not explicitly addressed in \cite{J. Huang 2023 TRO}. To handle this, the passed passage segment by the pivot path is translated to obstacle vertices to make \textit{translated passage segments}. As illustrated in Fig. \ref{Generalized Passage Segments}, $P_{\sigma_p}(1, i)$ is translated to vertices constructing this passage. $P_{\sigma_p}(1, i) = (\mathcal{E}_j, \mathcal{E}_k)$ with $\mathbf{p}_{j}^* \in \mathcal{E}_j, \mathbf{p}_{k}^* \in \mathcal{E}_k$ minimizing the distance in (\ref{Nearest Point Passage Definition}). There exists a translated passage segment starting at a vertex $\mathbf{v}^E_{j} \in \mathcal{E}_j$ if
\begin{equation}
\label{Valid Translated Passage Segment}
    \mathbf{v}^E_j + \delta \frac{ \mathbf{p}_k^* - \mathbf{p}_j^* }{ \| \mathbf{p}_k^* - \mathbf{p}_j^* \|_2 } \in \mathcal{X}_{free}
\end{equation}
for some small $\delta > 0$. $\mathbf{p}_k^* - \mathbf{p}_j^*$ characterizes the direction from $\mathcal{E}_j$ to $\mathcal{E}_k$ along $P_{\sigma_p}(1, i)$. For vertices on $\mathcal{E}_k$, the direction reverses. (\ref{Valid Translated Passage Segment}) essentially defines a ray starting at $\mathbf{v}^E_j$. The other end of the ray is assigned by detecting its collision with other obstacles or boundaries. Next, reference points of $\sigma_{t, i}$ on translated passage segments are found where the same procedure as in Section \ref{Path Set Generation Section} is adopted.
\begin{algorithm}[t]
    \nl $\Sigma_t \leftarrow \emptyset, \Sigma_S \leftarrow \emptyset$\;
    \nl $ \mathbf{s}_p \leftarrow$ (\ref{Pivot Selection})\;
    \nl $ \sigma_p \leftarrow$ {PAOPP in Algorithm (\ref{RRTStar})}\;
    \nl $\Sigma_t \leftarrow$ {(\ref{Combined Path Trasfer}) using $\sigma_p$}\;
    \nl \ForEach{$P_{\sigma_p}(1,i) \in P_{\sigma_p}(1)$} {
        \nl $\{\sigma_{t,1}(\eta_{1,i}), ..., \sigma_p(\eta_{p,i}), ..., \sigma_{t,K}(\eta_{K, i})\} \leftarrow \Sigma_t \cap P'_{\sigma_p}(1,i)$\;
        \nl $ \| \Sigma_t \cap P'_{\sigma_p}(1,i) \|_2 \leftarrow$ (\ref{Chord Length})\;
        \nl \If{$\| \Sigma_t \cap P'_{\sigma_p}(1,i) \|_2 > \| P_{\sigma_p}(1,i) \|_2$} {
            \nl $\sigma_p^*(\eta_{p,i}) \leftarrow$ (\ref{Reference Point Position 2})\;
        }    
        \nl \ElseIf{$(\Sigma_t \cap P'_{\sigma_p}(1,i)) \nsubseteq  P_{\sigma_p}(1,i)$} {
            \nl $\sigma_p^*(\eta_{p,i}) \leftarrow$ (\ref{Reference Point Position 1})\;
        }
        \nl \Else {
            \nl $\sigma_p^*(\eta_{p,i}) \leftarrow \sigma_p(\eta_{p,i})$\;
        }
    }    
    \nl $\{ \eta_{p,1}, \eta_{p,2}, ...\} \leftarrow \{0\} \cup \{\eta_{p,1}, \eta_{p,2}, ...\} \cup \{1\}$\;
    \nl \ForEach{$[\eta_{p,i}, \eta_{p,i+1}]$} {
        \nl $\sigma_p^*(\tau) \leftarrow$ (\ref{Reposition Pivot Path})\;
    }    
    \nl $\Sigma^*_t \leftarrow$ {(\ref{Combined Path Trasfer}) using $\sigma^*_p$}\;
    \nl \ForEach{$P_{\sigma_p}(1,i) \in P_{\sigma_p}(1)$} {
        \nl $\{\sigma_{t,1}^*(\eta_{1,i}), ..., \sigma_p^*(\eta_{p,i}), ..., \sigma_{t,K}^*(\eta_{K, i})\} \leftarrow \Sigma^*_t \cap P'_{\sigma_p}(1,i)$ and reference points generation\;
    }        
    \nl \ForEach{$\mathbf{s}_i \in S$ \textnormal{AND} $\mathbf{s}_i \neq \mathbf{s}_p$} {
        \nl $\sigma_{t,i}^* \leftarrow$ Reposition $\sigma_{t,i}$ as (\ref{Reposition Pivot Path})\; 
        \nl $\Sigma_S \leftarrow \Sigma_S \cup \{\sigma_{t,i}^*$\}\;
     }
    \nl \Return $\Sigma_S \cup \{ \sigma_p^* \}$\;
    \caption{Path Set Generation}
    \label{General Path Set Generation}
\end{algorithm}
\begin{figure*}[t]
    \minipage{2 \columnwidth}
    \centering
    \subfigure[$M = 10$] {
    \includegraphics[width= 0.31 \columnwidth]{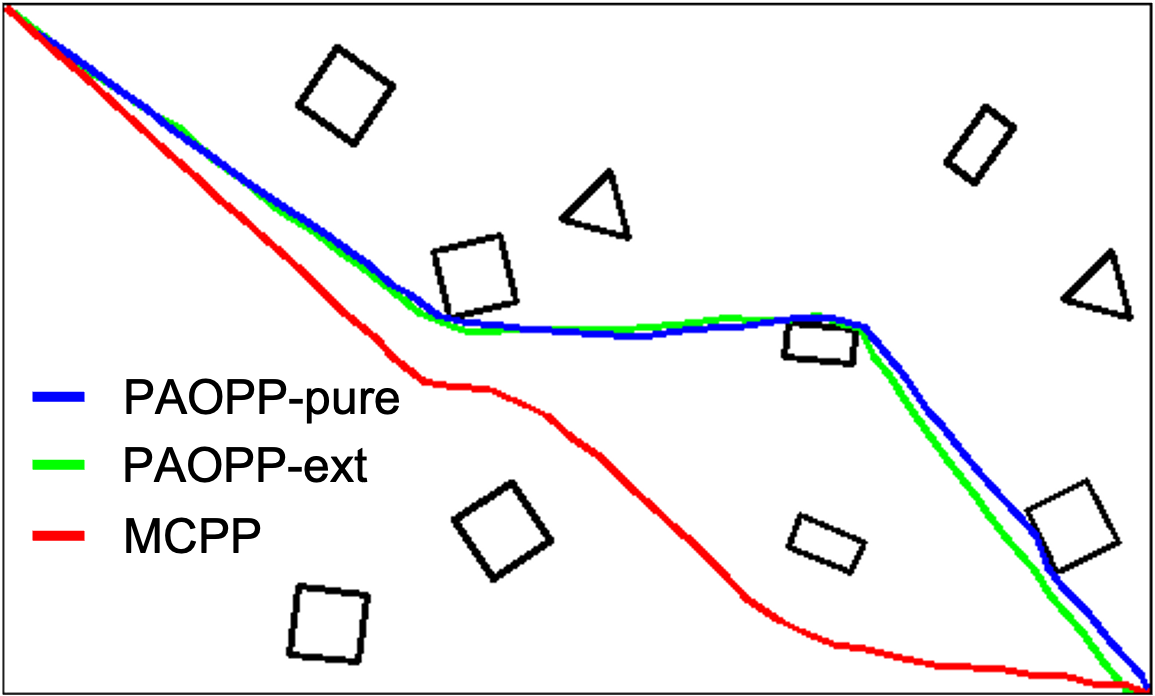}}
    \subfigure[{$M = 20$}] {
    \includegraphics[width= 0.31 \columnwidth]{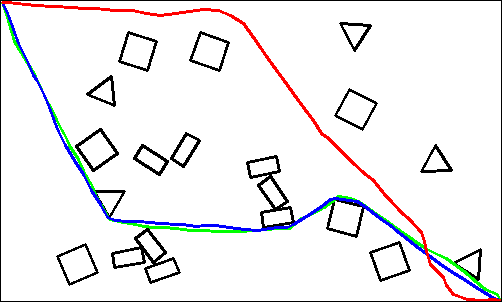}}
    \centering
    \subfigure[{$M = 30$}] {
    \includegraphics[width= 0.31 \columnwidth]{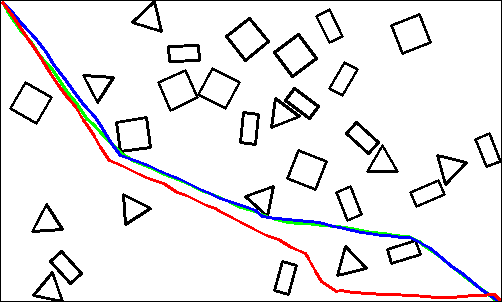}}
    \endminipage  \hfill 
    \minipage{2 \columnwidth}
    \centering
    \subfigure[{$M = 40$}] {
    \includegraphics[width= 0.31 \columnwidth]{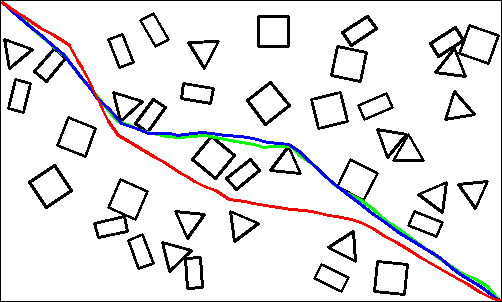}}
    \subfigure[{$M = 50$}] {
    \includegraphics[width= 0.31 \columnwidth]{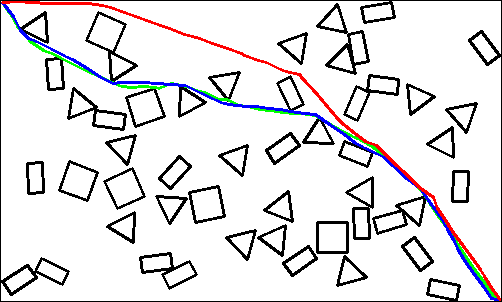}}
    \centering
    \subfigure[{$M = 60$}] {
    \includegraphics[width= 0.31 \columnwidth]{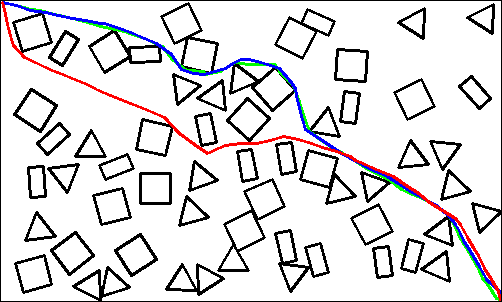}}
    \endminipage  \hfill     
    \caption{Path planning results using different planners. The obstacle number ranges from $10$ to $60$. Blue paths are found by PAOPP-pure. Green paths are found by PAOPP-ext. $k_P = 50$ in two PAOPP planners. Red paths are found by MCPP.} 
    \label{Path Planning With Different Planners}
\end{figure*}

\begin{figure}[t]
    \minipage{1 \columnwidth}
    \centering
    \includegraphics[width= 0.78 \columnwidth]{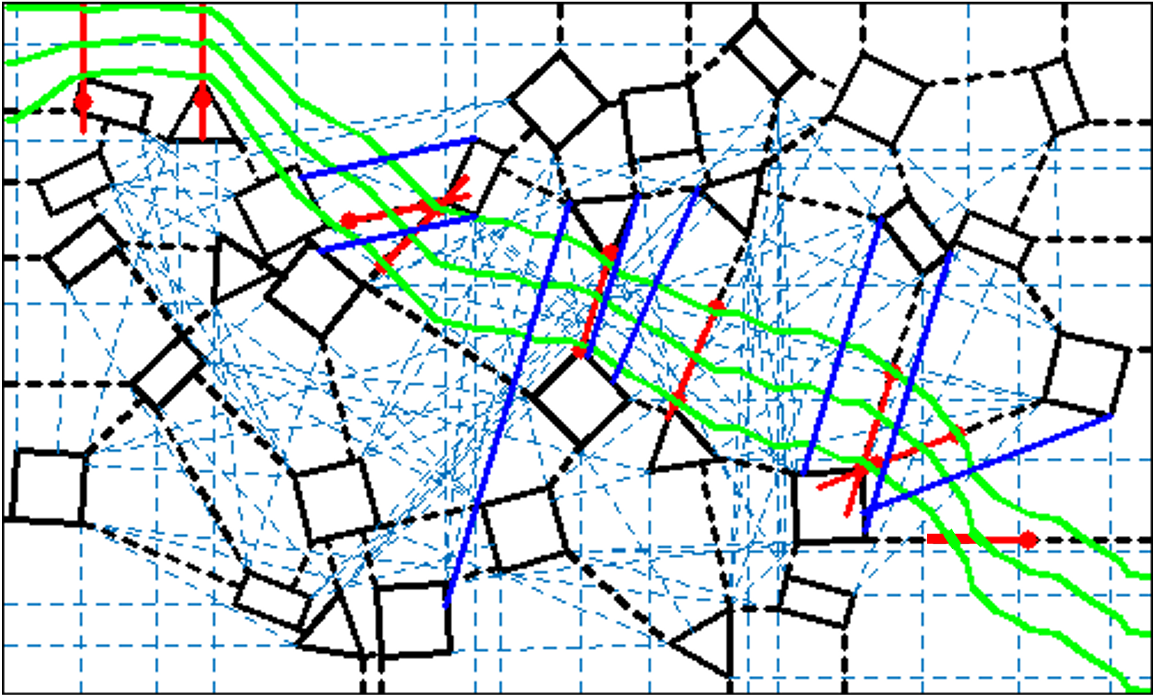}
    \endminipage \hfill
    \caption{Example of adding translated passages. Translated passage segments (blue segments) are only on obstacles on one side of the pivot path to avoid overly dense passage segments.}
    \label{Translated Passage Example}
\end{figure}

In practice, there may be multiple passage segments associated with an obstacle. For an obstacle vertex not contained in a passage, its nearest passage can be selected. Moreover, to avoid too dense passage segments, $P_{\sigma_p}(1, i)$ can be translated to only one obstacle in $\mathcal{E}_j$ and $\mathcal{E}_k$. Lastly, a translated passage can be filtered away by considering its distance to existing passages on the obstacle. Specifically, $d^*(\mathbf{v}^E_j, \mathbf{v}^E_{near})$ is the distance of $\mathbf{v}^E_j$ to an existing passage end $\mathbf{v}^E_{near}$ on $\mathcal{E}_j$ in the orthogonal direction of $\mathbf{p}_k^* - \mathbf{p}_j^*$, i.e.,
\begin{equation}
\label{Orthogonal Distance Between Generalized Passages}
    d^*(\mathbf{v}^E_j, \mathbf{v}^E_{near}) = \| (\mathbf{v}^E_{j} - \mathbf{v}^E_{near})\T  \mathbf{n}_{k,j}^{\bot} \|_2.
\end{equation}
$\mathbf{n}_{k,j}^{\bot}$ is the unit vector orthogonal to $\mathbf{p}^*_k - \mathbf{p}^*_j$. If $d^*(\mathbf{v}^E_j, \mathbf{v}^E_{near})$ is too small, $P_{\sigma_p}(1, i)$ is not translated to the vertex $\mathbf{v}^E_j$. An example of added passage segments is shown in Fig. \ref{Translated Passage Example}.

\section{Extended Experimental Results}
\label{App C}
More comparative results between PAOPP and MCPP, path set generation using SP and PT methods are reported here. Fig. \ref{Path Planning With Different Costs} shows examples of path planning results using MCPP and PAOPP with two passage detection strategies ($k_p = 50$). PAOPP-pure (using passages detected by the pure visibility check) and PAOPP-ext (using passages detected by the extended visibility check) find similar paths passing the same passages. Thus, paths have similar path costs and can be regarded as equivalently optimal paths. This empirically validates that the sparse passage distribution after the extended visibility check in PAOPP-ext will generally have the same optimal path as PAOPP-pure. The equivalence of PAOPP-ext and PAOPP-pure on an easy condition is analyzed in Appendix \ref{App D}. Paths found by MCPP may or may not be homotopic to those in PAOPP. Due to the clearance constraint, the planned path keeps the MC from obstacles. For better feasibility, the clearance to map boundaries is not included in MCPP. MCPP implementation is outlined in Algorithm \ref{Binary-Search MCPP}.

The comparisons between path set generation results using SP and PT are demonstrated in Fig. \ref{Path Set Generation Comparison}. In addition to the significant runtime differences reported in Table \ref{Tab: Path Set Planning Time}, Fig. \ref{Path Set Generation Comparison} provides intuitive comparisons of planned path sets. The main difference is that though the homotopy constraint is enforced in the SP method, coordination among paths is hard to achieve and paths overlap in most cases. To address this, explicit rules need to be specified to coordinate paths or postprocessing will be required to refine paths. In contrast, path sets generated by the PT method usually occupy wider spaces in narrow passages to place paths, which makes it more suitable for multiple path generation in obstacle-dense environments. In addition, the PT method allows for specifications of coordination requirements by adjusting reference point distributions in passages, making it quite flexible.
\begin{algorithm}[t]
    \nl Input $\mathbf{s}_{0}, \mathbf{s}_d$\;
    \nl $P_{valid} \leftarrow \texttt{PureVisibilityCheck}(\mathcal{E}_1, ..., \mathcal{E}_M)$\;
    \nl $MC_{lower} \leftarrow \min \| P_{valid} \|_2 / 2$\;
    \nl $MC_{upper} \leftarrow \max \| P_{valid} \|_2 / 2$\;
    \nl $\sigma_{MC} \leftarrow \emptyset$\;
    \nl \While{$ MC_{upper} - MC_{lower} > MC_{err}$} {
        \nl $MC \leftarrow MC_{lower} + (MC_{upper} - MC_{lower} ) / 2$\;
        \nl $\sigma_{MC} \leftarrow \texttt{PlanWithClearance}(\mathbf{s}_{0}, \mathbf{s}_d, MC)$\;
        \If{$\sigma_{MC} \neq \emptyset$} {
            \nl $MC_{lower} \leftarrow MC$\;
        }
        \Else {
            \nl $MC_{upper} \leftarrow MC$\;
        }
    }
    \nl \Return $\sigma_{MC}$\;
    \caption{Max-Clearance Path Planning (MCPP)}
    \label{Binary-Search MCPP}
\end{algorithm}

\section{Condition of Equivalence Between PAOPP-pure and PAOPP-ext}
\label{App D}
\begin{figure}[t]
    \minipage{1 \columnwidth}
     \centering
     \includegraphics[width= 0.8 \columnwidth]{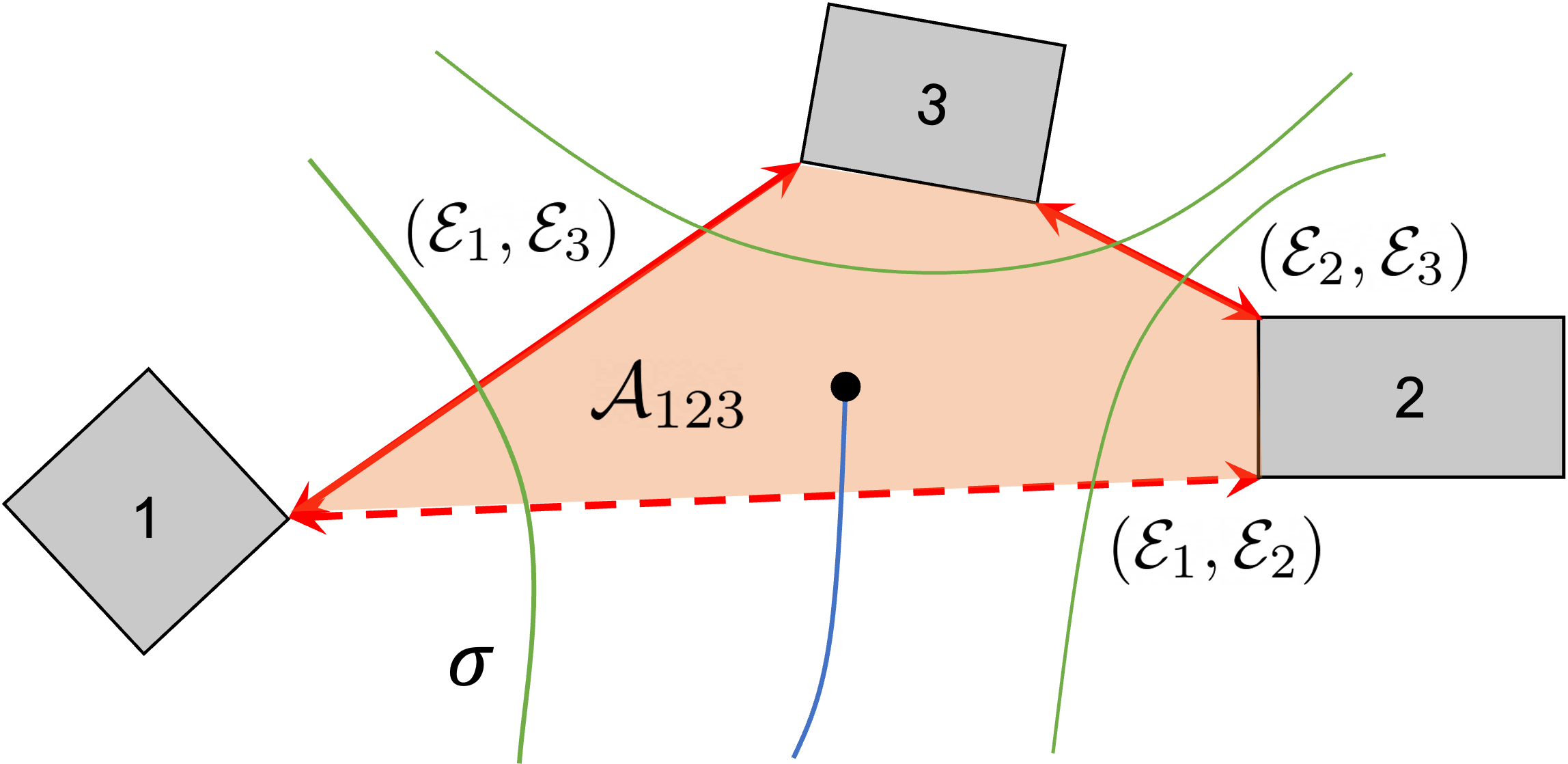}
     \endminipage \hfill
     \caption{Only type 3 and 4 are illustrated here. $(\mathcal{E}_1, \mathcal{E}_2)$ can pass the visibility check but cannot pass the extended variant. Green paths pass through $\mathcal{A}_{123}$ across different passages. The blue path has one endpoint inside $\mathcal{A}_{123}$.}
     \label{PAOPP Equivalence Example}
\end{figure}

One fundamental problem of the extended visibility check is whether it guarantees the optimality of planned paths in PAOPP problems. In other words, the problem is if paths planned by PAOPP-ext are guaranteed to be the same as those planned by PAOPP-pure. Here we show that the equivalence between PAOPP-ext and PAOPP-pure can be achieved via simple processing. For simplicity, only the 2D case is discussed here, but the derivation extends to 3D space readily. The elementary case is illustrated in Fig. \ref{PAOPP Equivalence Example}. The passage $(\mathcal{E}_1, \mathcal{E}_2)$ passes the pure visibility check but will be discarded in the extended visibility check due to $\mathcal{E}_3$. $(\mathcal{E}_1, \mathcal{E}_3)$ and $(\mathcal{E}_2, \mathcal{E}_3)$ are valid. Considering the region $\mathcal{A}_{123}$ enclosed by obstacles and passage segments, any non-winding path $\sigma$ can be categorized into one of the following four types:
\begin{enumerate}
    \item $\sigma$ does not pass $\mathcal{A}_{123}$.
    \item $\sigma$ is entirely inside $\mathcal{A}_{123}$.
    \item $\sigma$ passes through $\mathcal{A}_{123}$.
    \item One endpoint of $\sigma$ is inside $\mathcal{A}_{123}$ and the other is not.
\end{enumerate}

Our discussion is restricted to PAOPP problems in which $f_P(\sigma)$, the minimum passage width passed by $\sigma$, is used in the path cost like (\ref{Cost Function for Passage Passing}) and (\ref{Weighted Cost Function for Passage Passing}). For type 1 and 2, $f_P(\sigma)$ is intact. For type 3, the discarding of $(\mathcal{E}_1, \mathcal{E}_2)$ will not affect the update of $f_P(\sigma)$ before and after passing $\mathcal{A}_{123}$. Specifically, note that $\| (\mathcal{E}_1, \mathcal{E}_2) \|_2 > \| (\mathcal{E}_1, \mathcal{E}_3) \|_2$, $\| (\mathcal{E}_1, \mathcal{E}_2) \|_2 > \| (\mathcal{E}_2, \mathcal{E}_3) \|_2$. Then if $\sigma$ passes $(\mathcal{E}_1, \mathcal{E}_2)$ and $(\mathcal{E}_1, \mathcal{E}_3)$ (the passing order does not matter), the update rule of $f_P(\sigma)$ is 
\begin{equation}
    f_P(\sigma^2) = \min(f_P(\sigma^1), \, \| (\mathcal{E}_1, \, \mathcal{E}_3) \|_2)
\end{equation} 
where $\sigma^1, \sigma^2$ represent $\sigma$ before entering $\mathcal{A}_{123}$ and after leaving $\mathcal{A}_{123}$, respectively. If $\sigma$ passes $(\mathcal{E}_1, \mathcal{E}_2)$ and $(\mathcal{E}_2, \mathcal{E}_3)$, $f_P(\sigma)$ is updated as 
\begin{equation}
    f_P(\sigma^2) = \min(f_P(\sigma^1), \, \| (\mathcal{E}_2, \, \mathcal{E}_3) \|_2).
\end{equation} 
Finally, if $\sigma$ passes $(\mathcal{E}_1, \mathcal{E}_3)$ and $(\mathcal{E}_2, \mathcal{E}_3)$, $f_P(\sigma^2)$ is  
\begin{equation}
    f_P(\sigma^2) = \min(f_P(\sigma^1), \, \min(\| (\mathcal{E}_1, \, \mathcal{E}_3) \|_2, \|(\mathcal{E}_2, \, \mathcal{E}_3) \|_2).
\end{equation} 
In all cases, $f_P(\sigma^2)$ is independent from $(\mathcal{E}_1, \mathcal{E}_2)$. Therefore, discarding $(\mathcal{E}_1, \mathcal{E}_2)$ will not affect the update of the path cost $f(\sigma)$. This result holds iteratively for the entire environment. As a result, optimal planners return the path with the optimal cost, namely the optimal path.
\iftrue
\begin{figure*}[t]
    \minipage{2 \columnwidth}
    \centering
    \subfigure[SP ($M = 10, K = 3$)] {
    \includegraphics[width= 0.31\columnwidth]{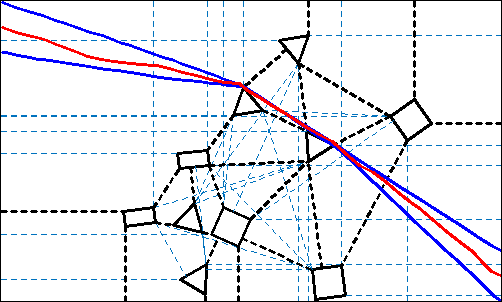}
    \label{SP 10 Obs 1}}
    \subfigure[SP ($M = 10, K = 9$)] {
    \includegraphics[width= 0.31\columnwidth]{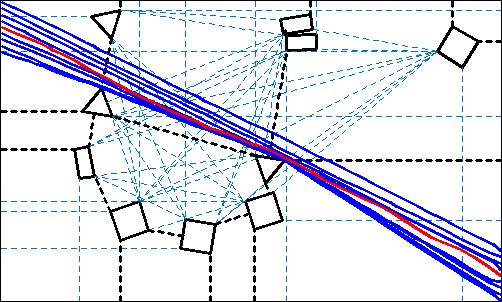}
    \label{SP 10 Obs 2}}
    \subfigure[SP ($M = 10, K = 15$)] {
    \includegraphics[width= 0.31\columnwidth]{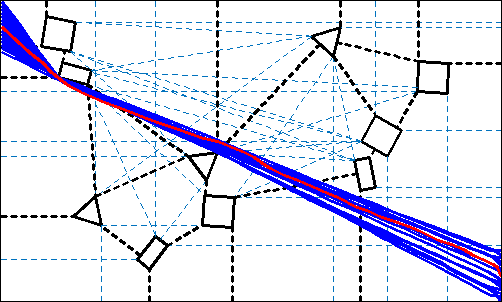}
    \label{SP 10 Obs 3}}
    \endminipage  \hfill 
    \minipage{2 \columnwidth}
    \centering
    \subfigure[PT ($M = 10, K = 3$)] {
    \includegraphics[width= 0.31\columnwidth]{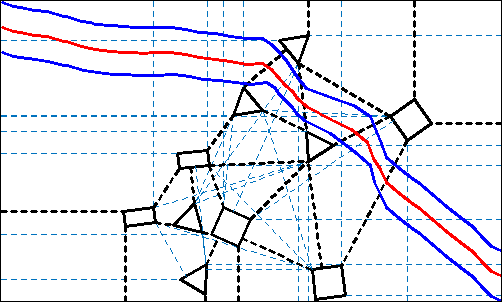}
    \label{PT 10 Obs 1}}
    \subfigure[PT ($M = 10, K = 9$)] {
    \includegraphics[width= 0.31\columnwidth]{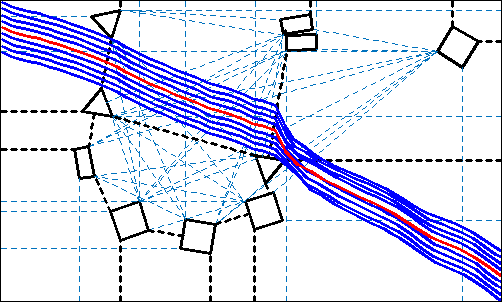}
    \label{PT 10 Obs 2}}
    \subfigure[PT ($M = 10, K = 15$)] {
    \includegraphics[width= 0.31\columnwidth]{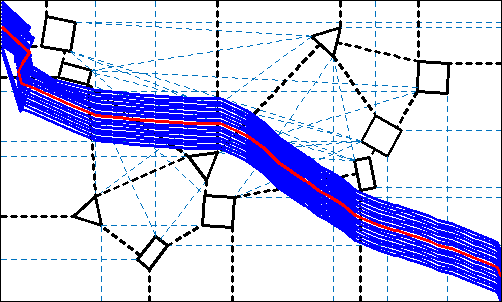}
    \label{PT 10 Obs 3}}
    \endminipage  \hfill     
    \minipage{2 \columnwidth}
    \centering
    \subfigure[SP ($M = 20, K = 3$)] {
    \includegraphics[width= 0.31\columnwidth]{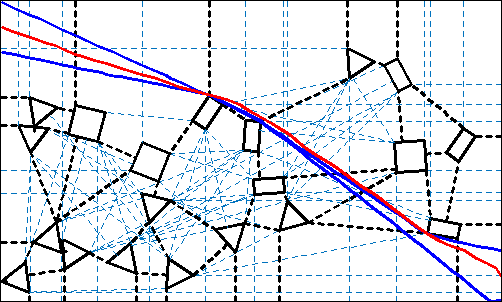}
    \label{SP 20 Obs 1}}
    \subfigure[SP ($M = 20, K = 9$)] {
    \includegraphics[width= 0.31\columnwidth]{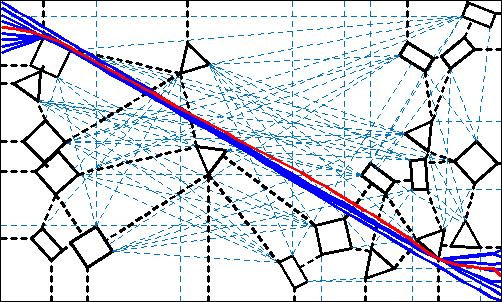}
    \label{SP 20 Obs 2}}
    \subfigure[SP ($M = 20, K = 15$)] {
    \includegraphics[width= 0.31\columnwidth]{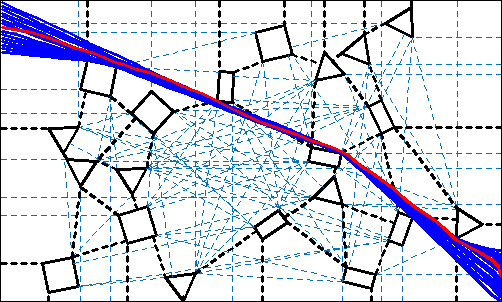}
    \label{SP 20 Obs 3}}
    \endminipage  \hfill 
    \minipage{2 \columnwidth}
    \centering
    \subfigure[PT ($M = 20, K = 3$)] {
    \includegraphics[width= 0.31\columnwidth]{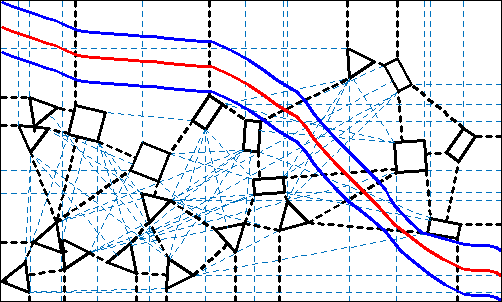}
    \label{PT 20 Obs 1}}
    \subfigure[PT ($M = 20, K = 9$)] {
    \includegraphics[width= 0.31\columnwidth]{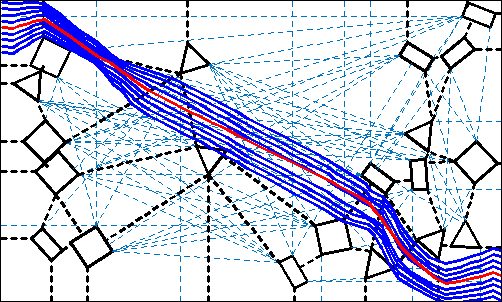}
    \label{PT 20 Obs 2}}
    \subfigure[PT ($M = 20, K = 15$)] {
    \includegraphics[width= 0.31\columnwidth]{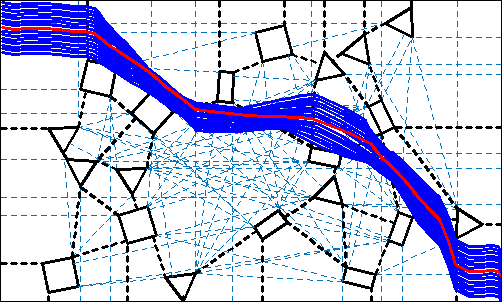}
    \label{PT 20 Obs 3}}
    \endminipage  \hfill    
    \minipage{2 \columnwidth}
    \centering
    \subfigure[SP ($M = 30, K = 3$)] {
    \includegraphics[width= 0.31\columnwidth]{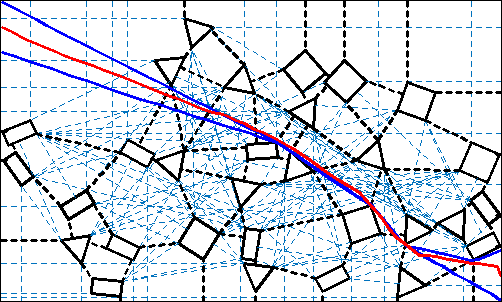}
    \label{SP 30 Obs 1}}
    \subfigure[SP ($M = 30, K = 9$)] {
    \includegraphics[width= 0.31\columnwidth]{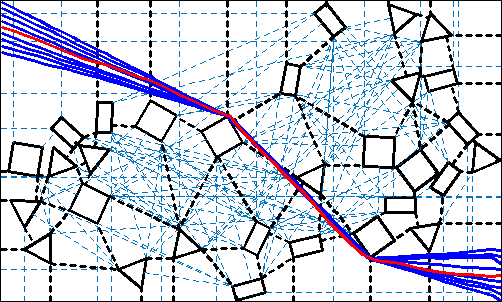}
    \label{SP 30 Obs 2}}
    \subfigure[SP ($M = 30, K = 15$)] {
    \includegraphics[width= 0.31\columnwidth]{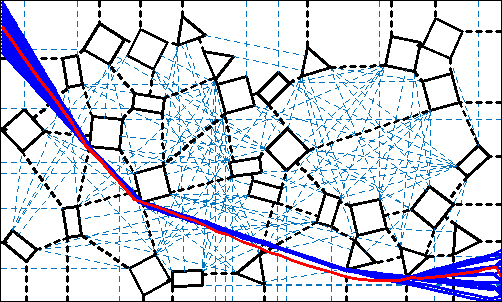}
    \label{SP 30 Obs 3}}
    \endminipage  \hfill 
    \minipage{2 \columnwidth}
    \centering
    \subfigure[PT ($M = 30, K = 3$)] {
    \includegraphics[width= 0.31\columnwidth]{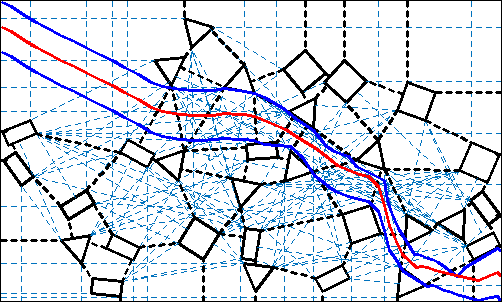}
    \label{PT 30 Obs 1}}
    \subfigure[PT ($M = 30, K = 9$)] {
    \includegraphics[width= 0.31\columnwidth]{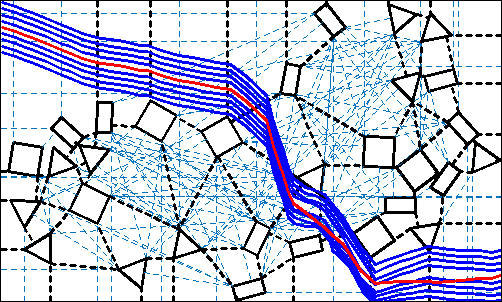}
    \label{PT 30 Obs 2}}
    \subfigure[PT ($M = 30, K = 15$)] {
    \includegraphics[width= 0.31\columnwidth]{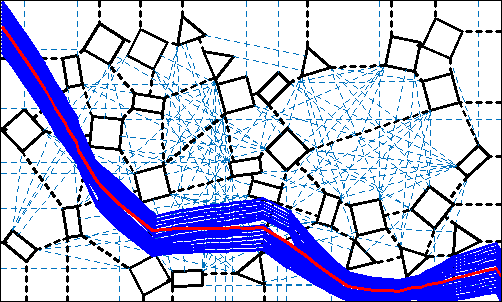}
    \label{PT 30 Obs 3}}   
        \endminipage  \hfill 
    \caption{Path set generation results in different setups using SP and PT methods. Red paths are pivot paths in SP and repositioned pivot paths in PT. Other agent paths are in blue.} 
    \label{Path Set Generation Comparison}
\end{figure*}
\fi

For type 4, however, PAOPP-ext and PAOPP-pure may return different results. The fundamental reason is that in PAOPP-ext, path costs of paths passing $(\mathcal{E}_1, \mathcal{E}_2)$ may not be updated in type 4. Assume the target position $\sigma(1)$ is within $\mathcal{A}_{123}$, in PAOPP-pure, $f_P(\sigma_{1,2}^2)$ after $\sigma$ passes $(\mathcal{E}_1, \mathcal{E}_2)$ is  
\begin{equation}
    f_P(\sigma_{1,2}^2) = \min(f_P(\sigma_{1,2}^1), \| (\mathcal{E}_1, \, \mathcal{E}_2) \|_2)
\end{equation} 
where the subscript of $\sigma$ signifies which passage $\sigma$ passes. In PAOPP-ext, however, $f_P(\sigma_{1,2}^2) = f_P(\sigma_{1,2}^1)$ is not updated. If $f_P(\sigma_{1,2}^1) > \| (\mathcal{E}_1, \, \mathcal{E}_2) \|_2$ happens, $f_P(\sigma_{1,2}^2) = \| (\mathcal{E}_1, \, \mathcal{E}_2) \|_2$ is correctly updated in PAOPP-pure, but $f_P(\sigma_{1,2}^2) = f_P(\sigma_{1,2}^1)$ in PAOPP-ext and a smaller cost is wrongly adopted (suppose other conditions except $f_P(\sigma)$ are the same). This may further lead to a different choice of the optimal path among $\sigma_{1,2}^2, \sigma_{1,3}^2$ and $\sigma_{2,3}^2$. The same also applies to situations where the start of $\sigma$ is in $\mathcal{A}_{123}$. Note that such a situation is rare and only locally affects the path's beginning or ending parts. A simple way to address this is to preserve passages enclosing the start and target positions in PAOPP-ext. To sum up, PAOPP-ext and PAOPP-pure are equivalent in finding the optimal path in type 1 to 3. In type 4, the equivalence can be ensured by preserving passages enclosing the start and target positions. 
\end{document}